\documentclass{article}
\pdfoutput=1

\usepackage{arxiv}

\usepackage[utf8]{inputenc} 
\usepackage[T1]{fontenc}    
\usepackage{hyperref}       
\usepackage{nicefrac}       
\usepackage{microtype}      
\usepackage[numbers]{natbib}
\usepackage{doi}

\usepackage{booktabs} 
\usepackage{amsmath, amssymb}
\usepackage[ruled, lined, onelanguage, linesnumbered]{algorithm2e}
\usepackage{graphics, float, url}
\usepackage{graphicx}
\usepackage{subcaption}
\usepackage{hyperref}
\usepackage{mathtools}
\usepackage{multirow}
\usepackage{adjustbox}
\usepackage{setspace}
\usepackage{amsfonts}
\usepackage{float}
\usepackage{lscape} 
\usepackage{pdflscape}
\usepackage{rotating}
\usepackage{tabularx}
\usepackage{color, colortbl}
\usepackage{stmaryrd}
\usepackage{bbold}
\usepackage{ifsym}
\usepackage{bbding}
\usepackage{fontawesome}

\title{Semi-supervised Clustering with Two Types of Background Knowledge: Fusing Pairwise Constraints and Monotonicity Constraints}

\author{
        \href{https://orcid.org/0000-0001-7881-2023}{\includegraphics[scale=0.06]{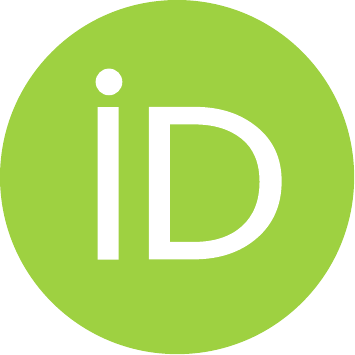}\hspace{1mm}Germ\'an Gonz\'alez-Almagro }\\
	DaSCI Andalusian Institute \\ DECSAI \\
	University of Granada\\
	Granada, Spain\\
	\texttt{germangalmagro@ugr.es} \\
        \And
        Juan Luis Su\'arez \\
	DaSCI Andalusian Institute \\ DECSAI \\
	University of Granada\\
	Granada, Spain\\
	\texttt{jlsuarezdiaz@ugr.es} \\
        \And
        Pablo S\'anchez-Bermejo \\
	DECSAI \\
	University of Granada\\
	Granada, Spain\\
	\texttt{pablo.snz.brm@gmail.com} \\
	\And
	Jos\'e-Ram\'on Cano \\
        DaSCI Andalusian Institute \\
	Dept. of Computer Science \\
        University of Ja\'en\\
	Ja\'en, Spain\\
	\texttt{jrcano@ujaen.es} \\
	\And
	Salvador Garc\'ia \\
	DaSCI Andalusian Institute \\ DECSAI \\
	University of Granada\\
	Granada, Spain\\
	\texttt{salvagl@decsai.ugr.es} \\
}

\renewcommand{\headeright}{}
\renewcommand{\undertitle}{}
\renewcommand{\shorttitle}{}

\hypersetup{
pdftitle={Semi-supervised Clustering with Two Types of Background Knowledge: Fusing Pairwise Constraints and Monotonicity Constraints},
pdfsubject={Computer Sciences, Data Science},
pdfauthor={Germ\'an Gonz\'alez-Almagro },
pdfkeywords={Pairwise constraints, Monotonicity constraints, Expectation-minimization, Semi-supervised learning, Machine learning},
}

\begin{document}
\maketitle

\begin{abstract}

This study addresses the problem of performing clustering in the presence of two types of background knowledge: pairwise constraints and monotonicity constraints. To achieve this, the formal framework to perform clustering under monotonicity constraints is, firstly, defined, resulting in a specific distance measure. Pairwise constraints are integrated afterwards by designing an objective function which combines the proposed distance measure and a pairwise constraint-based penalty term, in order to fuse both types of information. This objective function can be optimized with an EM optimization scheme. The proposed method serves as the first approach to the problem it addresses, as it is the first method designed to work with the two types of background knowledge mentioned above. Our proposal is tested in a variety of benchmark datasets and in a real-world case of study.

\end{abstract}

\keywords{Pairwise constraints \and Monotonicity constraints \and Expectation-minimization \and Semi-supervised learning \and Machine learning}

\section{Introduction}

Clustering constitutes a key research area in the unsupervised learning paradigm, where no information on how data should be handled is available. It can be viewed as the task of grouping instances from a dataset into groups (or clusters), with the aim to extract new information from them \cite{ezugwu2022comprehensive}. From the classic K-means algorithm to the newer proposals \cite{cai2023seeking}, clustering has been applied to many problems, such as time series monitoring \cite{enes2023pipeline}, COVID-19 medical image segmentation \cite{abd2021automatic} and regular image segmentation \cite{guo2023pixel}, noisy speech processing \cite{vani2019fuzzy} or band selection in hyperspectral images \cite{wang2022hyperspectral}. Background knowledge can be integrated into the classic clustering framework, thus reframing it into the semi-supervised learning paradigm \cite{van2020survey,chapelle2009semi}, where partial or incomplete information about the dataset is given to the perform clustering. 

When additional information is given in the form of constraints, the constrained clustering problem arises. Constraints can be understood in three main ways: cluster-level \cite{bradley2000constrained}, instance-level pairwise (or simply pairwise) \cite{davidson2007survey} and feature-level constrained clustering \cite{schmidt2011clustering}. This study focuses on pairwise constraints, which indicate whether two specific instances of a dataset must be placed in the same or in different clusters, resulting in Must-link (ML) and Cannot-link (CL) constraints, respectively. Constrained clustering has been applied in a variety of real world problems before, such as: satellite image time series \cite{lafabregue2019deep}, storage location assignment in warehouses \cite{yang2016constrained}, obstructive sleep apnea analysis \cite{mai2018evolutionary} or electoral district design \cite{balafar2020active}. Recent studies and proposals, such as \cite{gao2023towards}, prove the growing interest in the area of constrained clustering.

Recently, a new type of background knowledge coming from the supervised learning paradigm has been integrated into unsupervised learning. Monotonic classification is a particular case of supervised learning where classes are a set of ordered categories and classification models must respect monotonicity constraints among instances based on their descriptive features. This means that, if an instance $x_i$ has greater feature values than those of instance $x_j$, its assigned class must also be higher (greater) in the ordering than that of $x_j$ \cite{rosenfeld2021assessing}. Considering the classic example of house pricing: for two houses in the same neighborhood, the bigger ones are constrained to have higher prices than smaller houses when the rest of the features of the houses are similar \cite{gonzalez2021fuzzy}. This defines an order relationship between houses (instances) based on the value of their features, therefore models predicting house prices must take this into account to produce accurate results. Monotonicity constraints are a type of background knowledge that can be used to produce more accurate predictive models \cite{cano2019monotonic}, and has been successfully applied in real world problem such as fraudulent firm classification \cite{pan2020fraudulent}, real-time dynamic malware detection \cite{chistyakov2018monotonic}, or learning activities analysis based on students' opinion surveys \cite{cano2017prototype}. Additionally, a recent study from The Alan Turing Institute states that considering underlying data monotonicity in data science/machine learning models leads to fairer applications \cite{leslie2019understanding}.

In \cite{rosenfeld2021assessing} a methodology to perform clustering in the presence of monotonicity information (ordered clustering) is proposed within the Multi Criteria Decision Aid (MCDA) framework. It is done by defining a distance measure based on the concept of preference, which is later explained in detail (Section \ref{sec:PODC}), and whose basic concept relies on comparing instances from the dataset, discriminating feature-level comparative relationships. This results in a distance measure that produces ordered labeling in terms of monotonicity, understanding it as in the monotonic classification models which have previously been described.

This study addresses the fusion of the two types of background knowledge mentioned above: pairwise constraints and monotonicity constraints. It extends a previous study by the same authors \cite{gonzalez2022monotonic} in both the theoretical background of the proposed method and the testing of its capabilities. A real-world application is also presented in this study, addressing the Shanghai Ranking of World Universities (SRWU) dataset from a new perspective. A previous study which combines monotonicity constraints and cluster-size constraints (capacitated clustering) can be found in \cite{rosenfeld2021lexicographic}, where researchers are motivated by the existence of problems in which both types of background knowledge is available. This constitutes evidence in favor of the interest in the combination of different types of background knowledge, as there are real-world problems in which background knowledge is given in a heterogeneous fashion. Following this trend, our research is motivated by the existence of real-world problems in which monotonicity constraints and pairwise constraints are available, such as the  SRWU partitioning problem. To the best of our knowledge, there is no previous research on this topic, as models to perform ordered clustering have emerged very recently. In this study, the logical relationship between monotonic classification and ordered clustering is tackled, producing the monotonic clustering paradigm, in which pairwise constraints are later included, resulting in Monotonic Constrained Clustering (MCC). An expectation-minimization (EM) scheme is proposed to optimize a hybrid objective function which fuses both monotonicity and pairwise constraints. The proposed hybrid objective function is composed of a monotonic distance metric and a penalty term for pairwise constraints violations. The overall proposed optimization method for MCC is coined as Pairwise Constrained K-Means - Monotonic (PCKM-Mono).

The rest of this study is organized as follows: background concerning classic clustering, pairwise constrained clustering, monotonic classification and ordered clustering, which is presented in Section \ref{sec:Backgroud} and whose content is later used in Section \ref{sec:Proposal} to introduce the proposed MCC method. Once the experimental setup used to carry out our experiments is presented in Section \ref{sec:ExpSetup},  Sections \ref{sec:ExpResults} and \ref{sec:statisticalAnalysis} report and analyze the experimental results obtained by the proposed method. A real-world case of study is carried out in Section \ref{sec:ShanghaiRank}, where our proposal is used to perform clustering on the SRWU dataset and compare the results obtained by other methods in the same task. Lastly, our conclusions are discussed in Section \ref{sec:Conclusions}.

\section{Background} \label{sec:Backgroud}

As stated before, partitional clustering is the action of grouping instances of a dataset into $k$ clusters. A dataset $X = \{x_1, \cdots, x_n\}$ contains $n$ instances, each one described by $u$ features. The $i$th instance from $X$ is noted as $x_i = (x_{[i,1]}, \cdots, x_{[i,u]})$. The goal of a clustering algorithm is to assign a class label $l_i$ to each instance in $X$. The result is a list of labels $L = [l_1, \cdots, l_n]$, with $l_i \in \{1, \cdots, k\} \; \forall i \in \{1,\cdots, n\}$, that effectively splits $X$ into $k$ non-overlapping clusters $c_i$ to form a partition called $C = \{c_1, \cdots, c_K\}$. The label associated with a given cluster $c_i$ can be accessed as $l(c_i)$. The cluster membership of every instance is determined by the similarity of the instance to the rest of instances in the same cluster, and the dissimilarity to instances in other clusters. Many types of distance measurements can be used to determine pairwise similarities \cite{jain1999data}.

\subsection{Constrained Clustering} \label{sec:CC}

In real world applications, it is common to have some information about the analyzed datasets, even if this information is not given in the form of labels. In pairwise constrained clustering, a set of constraints is given to guide the clustering process. Constraints involve pairs of instances, indicating whether they must or must not belong to the same cluster; thus, two types of pairwise constraints can be formalized:

\begin{itemize}
	
	\item Must-link (ML) constraints $C_=(x_i,x_j)$: instances $x_i$ and $x_j$ from $X$ must be placed in the same cluster.
	
	\item Cannot-link (CL) constraints $C_{\neq}(x_i,x_j)$: instances $x_i$ and $x_j$ from $X$ cannot be assigned to the same cluster.
	
\end{itemize}

It is known that ML constraints are transitive, reflexive and symmetrical, and therefore they constitute an equivalence relationship. This is not the case for CL constraints; however, they can be chained to deduce new ML constraints \cite{wagstaff2001constrained}. Pairwise constraints can be enforced in two ways: hard \cite{wagstaff2001constrained} and soft \cite{law2004clustering} constraints. The former must necessarily be satisfied in the output partition of any algorithm which makes use of them, while the latter are interpreted as strong suggestions by the algorithm but can be only partially satisfied in the output partition.

In CC (Constrained Clustering), the goal is to find a partition (clustering) of $k$ clusters such that $C = \{c_1, \cdots, c_k\}$ of $X$, ideally satisfying all constraints (in hard CC) or as many constraints as possible (in soft CC). The classic clustering requirements also have to be observed: it must be fulfilled that the sum of instances in each cluster $c_i$ is equal to the number of instances in $X$, which has been defined as $n = |X| = \sum_{i = 1}^{k} |c_i|$.

\subsection{Monotonicity Constraints in Classification}

Monotonicity constraints were originally integrated into the supervised learning classification task, leading to monotonic classification. It can be viewed as a special case of standard classification where the classes constitute a set of ordered categories. Monotonic classification models must respect monotonicity constraints between the feature values of the instances and their class labels \cite{cano2019monotonic}.

Formally, monotonic classification aims to predict the class label $y_i$ from an instance $x_i$ with $y \in \mathcal{Y} = \{l_1, \cdots, l_m\}$. The categories in $\mathcal{Y}$ are arranged in an order relation  $\prec$ such as $l_1 \prec l_2 \prec \cdots \prec l_m$. In doing so, features and class labels are monotonically constrained by the problem background knowledge i.e. $x_i \succeq x_j \rightarrow f(x_i) \ge f(x_j)$ where $x_i \succeq x_j$ implies that all features in $x_i$ compare to features in $x_j$ with operator $\ge$, this is: $x_{i,q} \ge x_{j,q} \; \forall q \in \{1, \cdots, u\} $ \cite{kotlowski2012nonparametric}. This given relationship between instances referred as dominance. In this case $x_1$ dominates $x_2$. The goal of monotonic classification is to build a classifier that does not violate monotonicity constraints (pairwise dominance relationships). The result is a monotonic classifier \cite{cano2019monotonic}.

Much in the same way as it is done with constrained clustering methods, a distinction can be done in monotonic classifiers: soft monotonic models try to minimize the number of monotonic constrains violation, while hard monotonic models always produce monotonic predictions (never violate monotonic constraints) \cite{gonzalez2021fuzzy}.

\subsection{Partially Ordered Data Clustering in MCDA} \label{sec:PODC}

In \cite{rosenfeld2021assessing} the monotonicity constraints are integrated into unsupervised learning to produce the ordered clustering framework. Particularly, they are integrated into the MCDA paradigm, which is a subfield of operational research that concerns the structuring and solving decision problems including multiple criteria \cite{roy1996multicriteria}. To do so, the classic symmetrical notion of distance in pattern recognition is replaced with the asymmetrical notion of preference from the MCDA paradigm. The preference of an instance over another evaluates the global advantages of the former over the latter with respect to some preference criteria. The notion of preference can be seen as a decomposition of a distance measure, taking into account the sign of the differences. To cluster instances in an MCDA context, the similarity between every pair of instances is evaluated in terms of preferences taking all the other alternatives into account. With this in mind, two instances are similar if they are preferred to or by the same set of instances. To formalize these concepts, let us consider the weighted $L_1$ distance (for the maximization case and without loss of generality) as in Equation \ref{eq1}, which can be simplified as in Equation \ref{eq4}, with $w_d \in [0,1]$ being the weight assigned to the $d$th feature.

\begin{equation}
L_1(x_i,x_j) = \sum_{d=1}^u w_d | x_{[i,d]} - x_{[j,d]}|.
\label{eq1}
\end{equation}

\begin{equation}
L_1(x_i,x_j) = \sum_{d:x_{[i,d]} > x_{[j,d]}}^u w_d x_{[i,d]} - w_d x_{[j,d]} + \sum_{d:x_{[j,d]} > x_{[i,d]}}^u w_d x_{[j,d]} - w_d x_{[i,d]}.
\label{eq4}
\end{equation}

Subsequently, let us define the preference of $x_i$ over $x_j$ as in Equation \ref{eq2}. To put this into words, $r(x_i,x_j)$ quantifies the sum of differences between $x_i$ and $x_j$ limited to the features in which $x_i$ has higher (lower) values than $x_j$ for the maximization (minimization) case. Intuitively, the preference $r(x_i,x_j)$ indicates the cumulative quantified value of the advantage of $x_i$ over $x_j$. Please note that, as it has already been mentioned, the preference is not symmetrical: $r(x_i,x_j) \neq r(x_j,x_i)$ in most cases. 

\begin{equation}
r(x_i,x_j) = \sum_{d:x_{[i,d]} > x_{[j,d]}}^u w_d x_{[i,d]} - w_d x_{[j,d]}.
\label{eq2}
\end{equation}

Finally, note that the weighted $L_1$ distance between two instances can always be expressed as in Equation \ref{eq3}. This decomposition can be done the same way for any $L_p$ distance.

\begin{equation}
L_1(x_i,x_j) = r(x_i,x_j) + r(x_j,x_i).
\label{eq3}
\end{equation}

\section{The Proposal: Pairwise Constrained Monotonic Clustering} \label{sec:Proposal}

In this study, the combination of pairwise constraints and monotonicity constraints is investigated. Bearing in mind all formal concepts from monotonic classification and ordered clustering (from Section \ref{sec:Backgroud}), establishing a logical relation between the concepts of dominance and preference is straightforward. This is: if an instance $x_i$ dominates $x_j$, then it is also true that instance $x_i$ is preferred over $x_j$ (for uniform weights). More formally: $x_i \succeq x_j \rightarrow r(x_i,x_j) \ge r(x_j,x_i)$. This way, any distance $L_p$ defined as in Equation \ref{eq3} can be used to measure distances in clustering methods for them to produce output partition satisfying monotonicity constraints. This new clustering paradigm is coined as monotonic clustering.

To perform pairwise constrained monotonic clustering, an Expectation-Minimization (EM) optimization scheme is used, along with a hybrid objective function which takes into account both pairwise constraints and monotonicity constraints. To this end, a distance measure designed on the basis of the definition of preference (originally used in ordered clustering), and a pairwise constraint-based penalty term are combined to produce the already mentioned function. We named this approach Pairwise Constrained K-Means - Monotonic (PCKM-Mono).

The EM optimization scheme is widely used in the literature to approach clustering problems ranging from classic clustering problems to constrained clustering \cite{vouros2021semi} and monotonic clustering \cite{rosenfeld2021assessing}. Two steps build the EM optimization scheme: (1) in the Expectation step (E step), given a set of cluster representatives (centroids) $\{\mu_1, \cdots, \mu_K\}$, every instance $x_i$ is assigned to the cluster $c_j$ that minimizes its contribution to the objective function, computed with respect to the cluster representatives; (2) in the Minimization step (M step), the cluster representatives $\{\mu_1, \cdots, \mu_K\}$ are reestimated for the current cluster assignment $\{c_1, \cdots, c_K\}$ to minimize the objective function. The EM optimization scheme iterates between these two steps until some convergence criteria are met. With this in mind, two elements need to be defined in order to apply the EM scheme to the constrained monotonic problem: the objective function and the centroid computation criteria.

\paragraph{Cost function.} The cost function of the proposed PCKM-Mono algorithm combines two main elements: a monotonic distance measure (proposed in \cite{rosenfeld2021assessing}) and a pairwise constraint-based penalty term. Equation \ref{eq:ObjectiveFunction} defines the hybrid objective function optimized by PCKM-Mono, where $\mathbb{1} \llbracket \cdot \rrbracket$ is the indicator function (returns 1 if the predicate given as argument holds, and 0 otherwise), and $\mu_k$ is the centroid associated with cluster $k$. The first term in Equation \ref{eq:ObjectiveFunction} is a preference-based distance metric, while the other two terms refer to the cost of violating CL and ML constraints (the penalty term), respectively. Please note that the first term of Equation \ref{eq:ObjectiveFunction} produces completely stratified clusters when applied alone, which would produce perfectly monotonic partitions. However, this is not a desirable result in most real-world problems, as will be proved in Section \ref{sec:ExpResults}.

\begin{equation}
\begin{array}{lc}
J_{PCKMM} = & \frac{1}{K}\sum_{k=1}^{K} \sum_{x_i \in c_k}|(r(x_i, \mu_k) - r(\mu_k, x_i))| + \\ &
\\ & \sum_{(x_i, x_j) \in C_=} \mathbb{1} \llbracket l_i \neq l_j \rrbracket + \sum_{(x_i, x_j) \in C_{\neq}} \mathbb{1} \llbracket l_i = l_j \rrbracket
\end{array}.
\label{eq:ObjectiveFunction}
\end{equation}

This cost function can be translated into an assignation rule as in Equation \ref{eq:AssignationRule}, which can be intuitively interpreted as: assign each instance to its closest (preferred) cluster among those where it produces the least violated constraints.

\begin{equation}
\begin{array}{lc}
x_i \in c_{h^*} \;\; \text{\textbf{if}} \;\; h^* =  & \texttt{argmin}_h \left( |\sum_{j=1}^{u}(x_{[i,j]}-\mu_{[h,j]})| + \right. \\
& \left. \sum_{x_j:(x_i, x_j) \in C_=} \mathbb{1} \llbracket l(c_h) \neq l_j \rrbracket + \sum_{x_j:(x_i, x_j) \in C_{\neq}} \mathbb{1} \llbracket l(c_h) = l_j \rrbracket \right)
\end{array}.
\label{eq:AssignationRule}
\end{equation}

\paragraph{Centroid update rule.} Regarding the computation of the centroid for every cluster after the E step, it is done by following its traditional form: every centroid is computed as the average of all instances which belong to the cluster it represents. This can be formalized as in Equation \ref{eq:UpdateRule}.

\begin{equation}
\mu_i = \frac{1}{|c_i|} \sum_{x_i \in c_i} x_i
\label{eq:UpdateRule}
\end{equation}

The overall PCKM-Mono optimization procedure is summarized in Algorithm \ref{alg:PCKMM}. It is clear that the proposed method is soft constrained for both pairwise constraints and monotonicity constraints.

\begin{algorithm}
	\SetNlSty{textbf}{[}{]}
	\SetNlSkip{0.5em}
	\SetKwRepeat{Do}{do}{while}
	\KwIn{Dataset $X$, constraint sets $C_=$ and $C_{\neq}$, the number of clusters $K$.}
	\KwOut{Partition $C$ of $K$ non-overlapping clusters.}
	\BlankLine
	
	Initialize centroids $\{\mu_1, \cdots, \mu_K\}$ randomly\\
	\Do{not converged}{
		\tcp{Expectation Step}
		\For{$i \in \{i,\cdots, n \}$}{
		    Assign each instance $x_i$ to cluster $h^*$ following Equation \ref{eq:AssignationRule}.
		}
		\tcp{Minimization Step}
		\For{$i \in \{i,\cdots, K \}$}{
		    $\mu_i = \frac{1}{|c_i|} \sum_{x_i \in c_i} x_i$
		}
		
	}
	\KwRet{$C$}
	
	\caption{Pairwise Constrained K-Means - Monotonic (PCKM-Mono)}\label{alg:PCKMM}
\end{algorithm}

\section{Experimental Setup and Calibration} \label{sec:ExpSetup}

In order to evaluate the capabilities of our proposal and compare its performance with previous methods, monotonic datasets need to be used. In \cite{gonzalez2021fuzzy} a list 12 monotonic datasets is used to test the capabilities of monotonic methods. These are the datasets used in our experiments, which can be found in the \href{https://sci2s.ugr.es/keel/category.php?cat=clas}{Keel-dataset repository}\footnote{https://sci2s.ugr.es/keel/category.php?cat=clas} \cite{triguero2017keel} and are used in recent research concerning monotonic classification \cite{zhu2021fuzzy}. Three constraint sets with incremental levels of constraint-based information are generated for each dataset. Since the Euclidean distance is used to measure pairwise distances in all compared algorithms, a standardization procedure is applied to all datasets. No other preprocessing step is performed on the datasets.

Constraints are generated following the method in \cite{wagstaff2001constrained}. Three constraint sets are generated for every datasets, namely: $CS_{10}$, $CS_{15}$ and $CS_{20}$. Each constraint set is associated with a small percentage of the size of the dataset: 10\%, 15\% and 20\%, respectively. The formula $(n_f(n_f-1))/2$ tells us how many artificial constraints will be created for each constraint set, with $n_f$ being the fraction of the size of the dataset associated with each of these percentages. Table \ref{tab:DatasetConts} displays a summary of all datasets and constraint sets used in our experiments.

\begin{table}[!h]
	\centering
	\setlength{\tabcolsep}{7pt}
	\renewcommand{\arraystretch}{1.1}
	\caption{Datasets and Constraint Sets Summary}
	\resizebox{1\textwidth}{!}{
		\begin{tabular}{lccc cc c cc c cc}
			\hline
			Dataset & Instances & Classes & Features &
			\multicolumn{2}{c}{$CS_{10}$} && \multicolumn{2}{c}{$CS_{15}$} && \multicolumn{2}{c}{$CS_{20}$} \\
			\cline{5-6} \cline{8-9} \cline{11-12}
			&&&& ML & CL && ML & CL && ML & CL \\
			\hline
            Artiset & 899 & 10 & 2 & 494 & 3422 && 1240 & 7671 && 2061 & 13870 \\
            Balance & 625 & 3 & 4 & 832 & 1059 && 1799 & 2479 && 3332 & 4418 \\
            BostonHousing4CL & 506 & 4 & 13 & 284 & 941 && 686 & 2089 && 1266 & 3784 \\
            Car & 1728 & 4 & 6 & 7961 & 6745 && 18167 & 15244 && 32076 & 27264 \\
            ERA & 1000 & 9 & 4 & 676 & 4274 && 1562 & 9613 && 2760 & 17140 \\
            ESL & 488 & 9 & 4 & 216 & 912 && 521 & 2107 && 949 & 3707 \\
            LEV & 1000 & 5 & 4 & 1381 & 3569 && 3174 & 8001 && 5692 & 14208 \\
            MachineCPU & 209 & 4 & 6 & 41 & 149 && 99 & 366 && 205 & 615 \\
            Qualitative Bankruptcy & 250 & 2 & 6 & 35 & 147 & &153 & 344 & &322 & 617 \\
            SWD & 1000 & 4 & 10 & 1566 & 3384 && 3674 & 7501 && 6583 & 13317 \\
            Windsor Housing & 546 & 2 & 11 & 915 & 516 && 2105 & 1135 && 3827 & 2059 \\
            Wisconsin & 683 & 2 & 9 & 1273 & 1005 && 2834 & 2317 && 5146 & 4034 \\
			\hline
			
	\end{tabular}}
	\label{tab:DatasetConts}
\end{table}

\subsection{Evaluation Method and Validation of Results} \label{sec:EvalValid}

Given the hybrid nature of our proposal, different features of the obtained partitions results have to be inspected to assess their quality in terms of different measures. The Adjusted Rand Index (ARI) will be used to measure the overall degree of agreement between the obtained partitions and the ground truth \cite{hubert1985comparing}. The Rand Index measures the degree of agreement of two partitions $C_1$ and $C_2$ for the same given dataset $X$, with $C_1$ and $C_2$ viewed as collections of $n(n - 1)/2$ pairwise decisions. This measure is corrected for chance to obtain the ARI. For more details on ARI, see \cite{hubert1985comparing}. An ARI value of 1 indicates total agreement between $C_1$ and $C_2$, while -1 means total disagreement. The quality with respect to the monotonicity of the obtained partition can be measured with the Non-Monotonic Index (NMI), which measures the degree to which monotonicity constraints are violated. It is defined as  the rate of violations of monotonicity divided by the total number of examples in a dataset \cite{gonzalez2015monotonic}. Finally, the \textit{Unsat} measure is used to evaluate the quality of the results from the point of view of constrained clustering. \textit{Unsat} is computed as the rate of violated constraints in a given partition \cite{gonzalez2020dils}.

Bayesian statistical tests are used in order to validate the results (which will be presented in Section \ref{sec:ExpResults}), instead of using the classic Null Hypothesis Statistical Tests (NHST), whose disadvantages are analyzed in \cite{benavoli2017time}, where a new statistical comparative framework is also proposed. The Bayesian version of the frequentist non-parametric sign test is used in this study. In the Bayesian sign test, the statistical distribution of a given parameter $\rho$ is obtained according to the differences between two sets of results, assuming it is a Dirichlet distribution. To do so, the Bayesian sign test proceeds as follows: the number of times that $A - B < 0$, the number of times where there are no significant differences, and the number of times that $A - B > 0$, then the weights of the Dirichlet distribution are iteratively updated and finally sampled to obtain a large sample of the distribution. In order to identify cases where there are no significant differences, the region of practical equivalence (rope) $[r_\text{min}, r_\text{max}]$ is defined, so that $P(A \approx B) = P(\rho \in \text{rope})$. The result of this process is a set of triplets with the form described in Equation \ref{eq:BST_triplet}. The \texttt{rNPBST} R package is employed to apply the test, whose documentation and guide can be found in \cite{carrasco2017rnpbst}.

\begin{equation}
\begin{array}{lc}
[P(\rho < r_\text{min}) = P(A - B < 0), \;\; P(\rho \in \text{rope}) 
P(\rho > r_\text{max}) = P(A - B > 0)]
\end{array}.
\label{eq:BST_triplet}
\end{equation}

\subsection{Calibration}

To demonstrate the capabilities of the proposed PCKM-Mono algorithm, it is compared with four other previous EM-style clustering algorithms, including the only existing purely monotonic clustering algorithm, two purely constrained clustering algorithms (including the most recent one), and a classic clustering algorithm:

\begin{itemize}

    \item P2Clust: The first approach to monotonic clustering. It modifies the distance measure used in the expectation step of the EM scheme to produce purely monotonic partitions. Monotonicity constraints are never violated in partitions produced by P2Clust \cite{rosenfeld2021assessing}, thus it is a hard constrained method for monotonicity constraints. It does not consider pairwise constraints, therefore it is purely monotonic.
    
    \item COP-Kmeans: COnstrained Partitional K-means constitutes the first approach to constrained clustering \cite{wagstaff2001constrained}. It is taken as the baseline comparison for any constrained clustering method. To integrate constraints into the clustering process, it modifies the assignment rule of instances to a cluster in such a way that no constraints can be violated. The algorithm halts when a dead-end is reached. It produces partitions which satisfy all constraints when it does not arrive at dead-ends, thus it is a hard constrained method for pairwise constraints. It a purely constrained clustering algorithm.
    
    \item Kmeans: The original Kmeans algorithm proposed in \cite{lloyd1982least}. Neither pairwise constraints nor monotonicity constraints are considered in Kmeans.
    
    \item PCSKMeans: The Pairwise Constrained Sparse K-Means algorithm is an extension of the classic Sparse K-Means algorithm that integrates constraints by means of a weighted penalty term \cite{vouros2021semi}. It constitutes the most recent EM-style approach to constrained clustering.
    
\end{itemize}

Regarding the parameter setup, all algorithms use an EM scheme to find a partition of the datasets, thus sharing many of their parameters. The $k$ parameter, which indicates the number of clusters of the output partition is always set to the number of classes for every dataset (in Table \ref{tab:DatasetConts}). The maximum number of iterations allowed before convergence is set to 100 in all cases. The convergence criterion is centroid shifting: the EM optimization procedure is considered to have converged when average centroid shifting is less than $10^{-4}$. Random centroid initialization is used for all algorithms. The P2Clust algorithm allows us to parameterize the computation of its internal $\alpha$ coefficient; this parameter is set to 1.1. The sparsity level of the PCSKMeans algorithm is set to 1.1. All parameters have been set by following the guidelines of the authors, and PCKM-Mono parameters have been decided upon preliminary experimentation. The final purpose of this work is to provide a fair comparison between algorithms, assessing their robustness in a common environment with multiple datasets.

\section{Experimental Results} \label{sec:ExpResults}

The experimental results obtained for all datasets and constraint sets are presented in this section. Since non-deterministic procedures are present in every compared method (such as the random initialization of centroids), the average results of 50 runs are presented in Tables \ref{tab:res10}, \ref{tab:res15} and \ref{tab:res20}, aiming to mitigate the effects that stochastic procedures may cause. Please note that, in cases where the COP-Kmeans algorithm is not able to produce a partition, we assign that particular run the worst possible benchmark values. Cases where no result is reported are cases in which COP-Kmeans was never able to produce an output partition. Let us remember that ARI is a maximization external quality index, while NMI and Unsat are both for minimization.

\begin{figure}[!h]
	\centering
	 \subfloat[]{\includegraphics[width=0.4\linewidth]{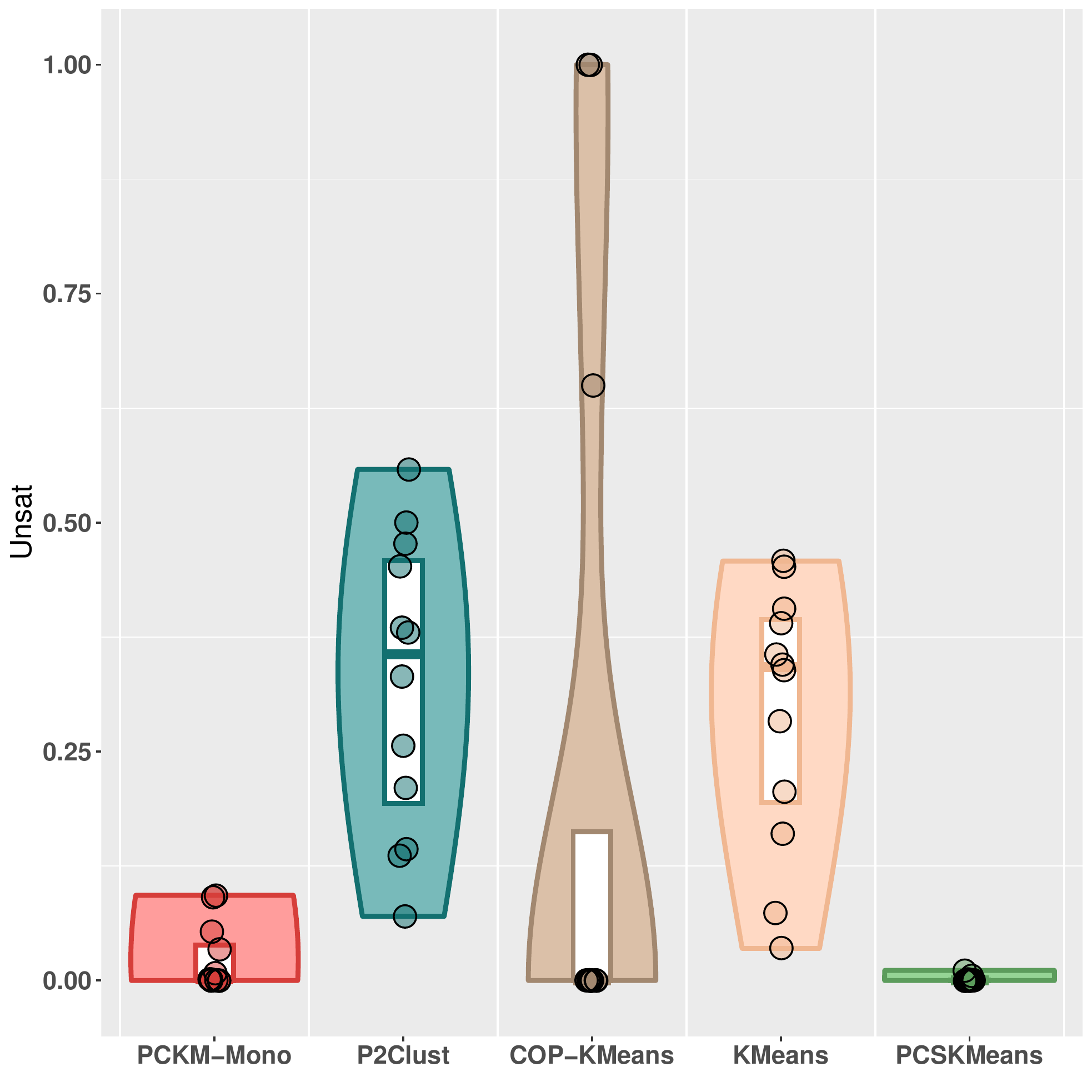}
		\label{fig:violin_unsat_10}}
	\subfloat[]{\includegraphics[width=0.4\linewidth]{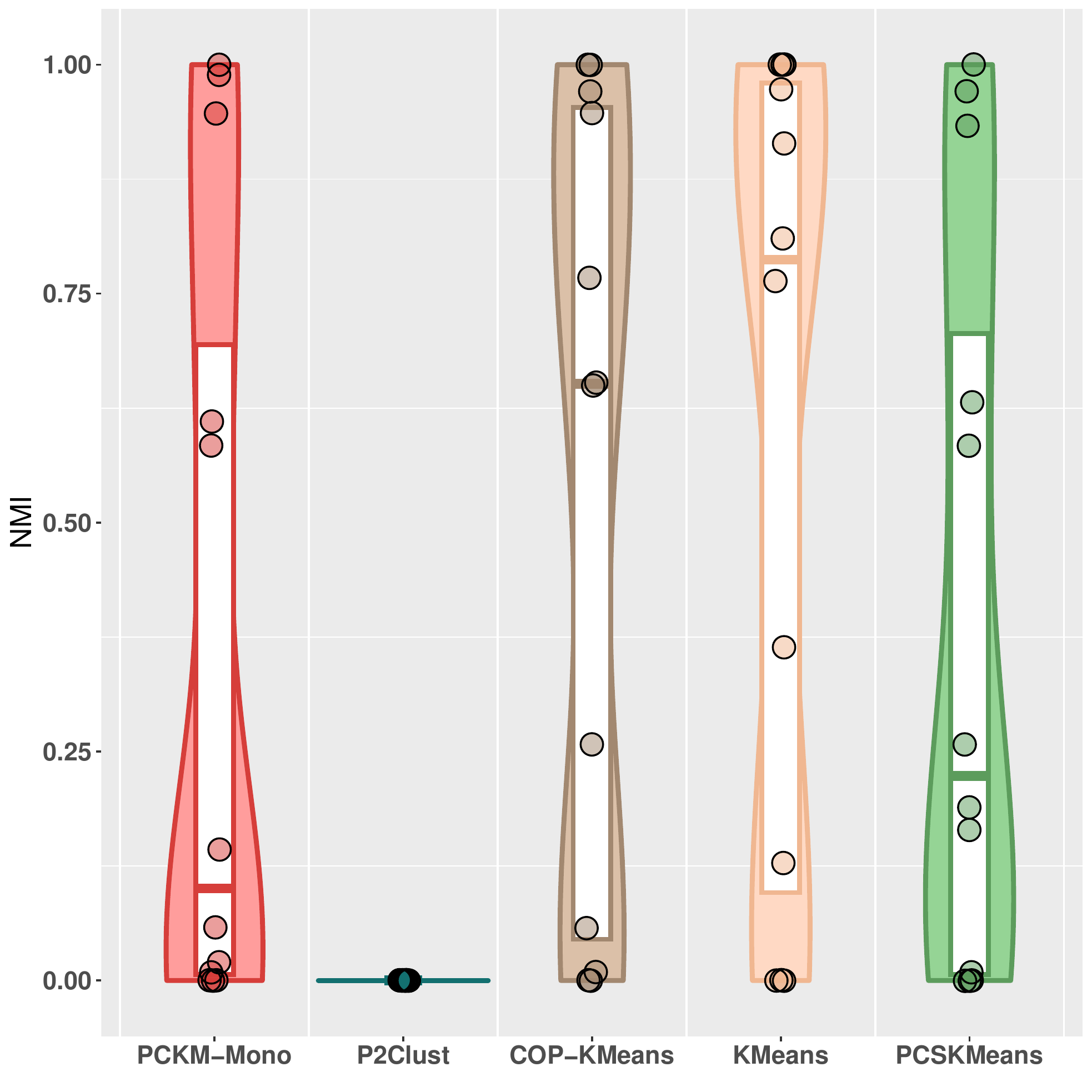}
		\label{fig:violin_nmi_10}}

    \vspace{\baselineskip}
  
	\subfloat[]{\includegraphics[width=0.8\linewidth]{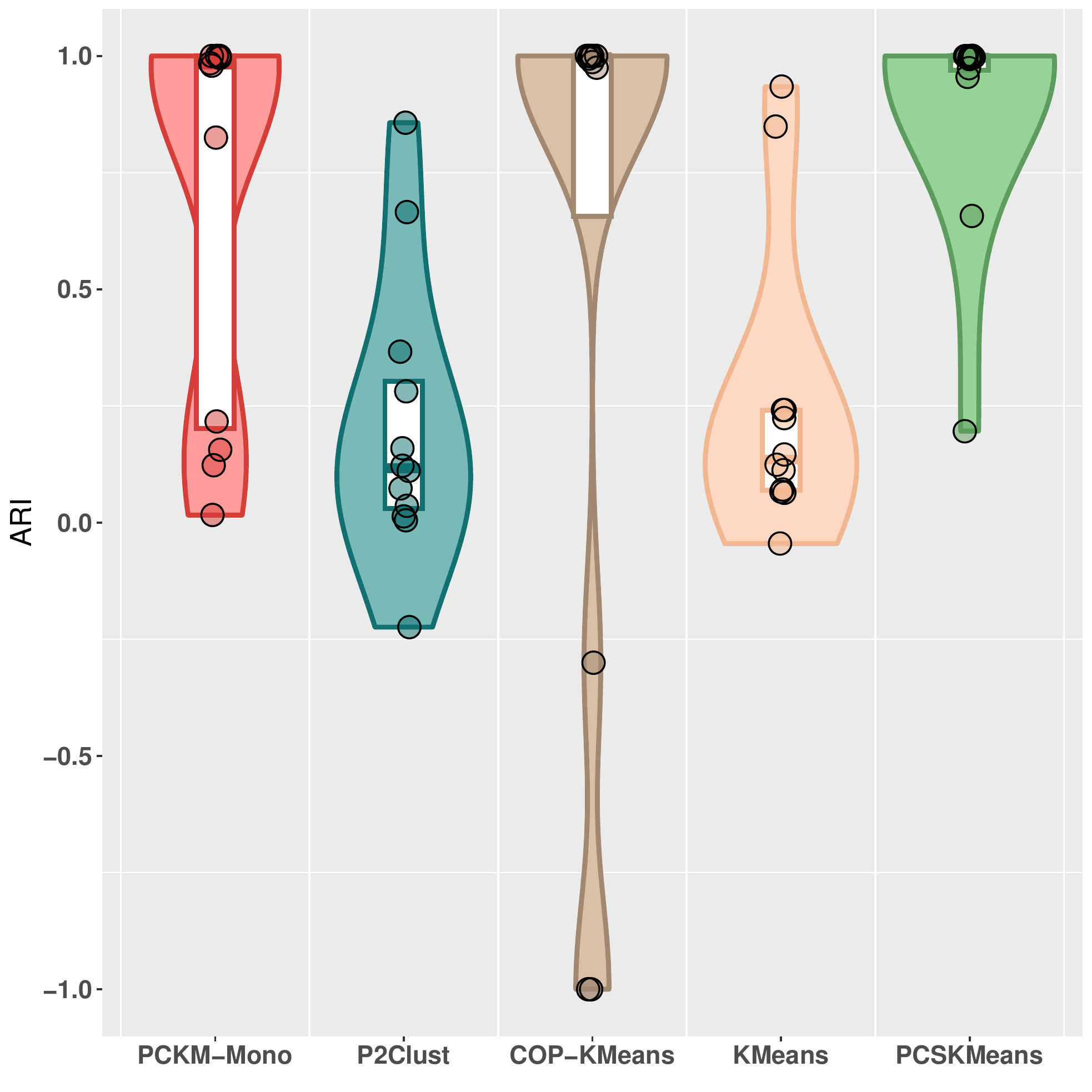}
		\label{fig:violin_ari_10}}

	\caption{Comparative violinplots for the results obtained with CS$_{10}$. Figures \ref{fig:violin_unsat_10}, \ref{fig:violin_nmi_10}, and \ref{fig:violin_ari_10} show the results obtained for the Unsat, NMI, and ARI measures, respectively.}
	\label{fig:ViolinPlots_CS10}
\end{figure}

\begin{figure}[!h]
	\centering
	 \subfloat[]{\includegraphics[width=0.4\linewidth]{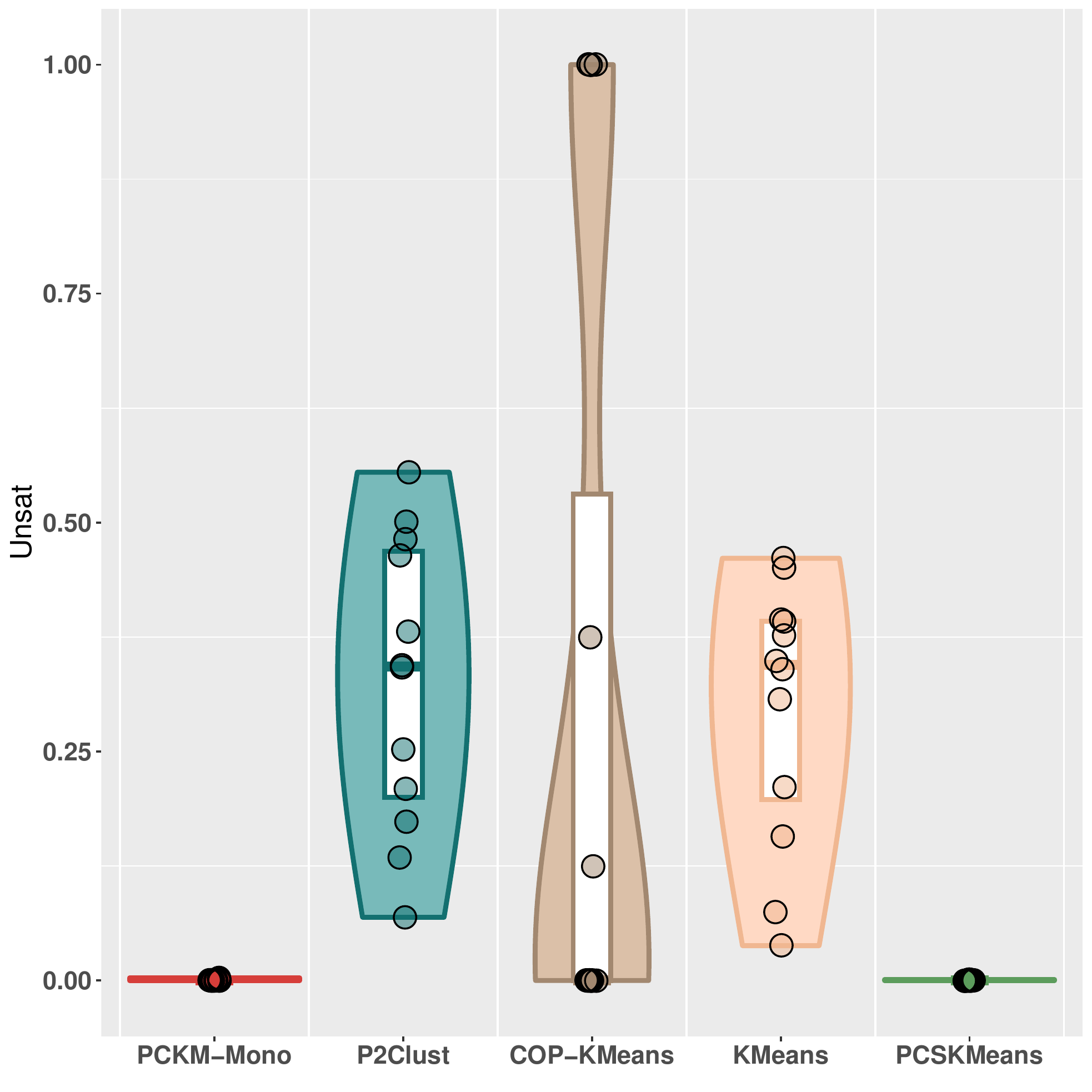}
		\label{fig:violin_unsat_15}}
	\subfloat[]{\includegraphics[width=0.4\linewidth]{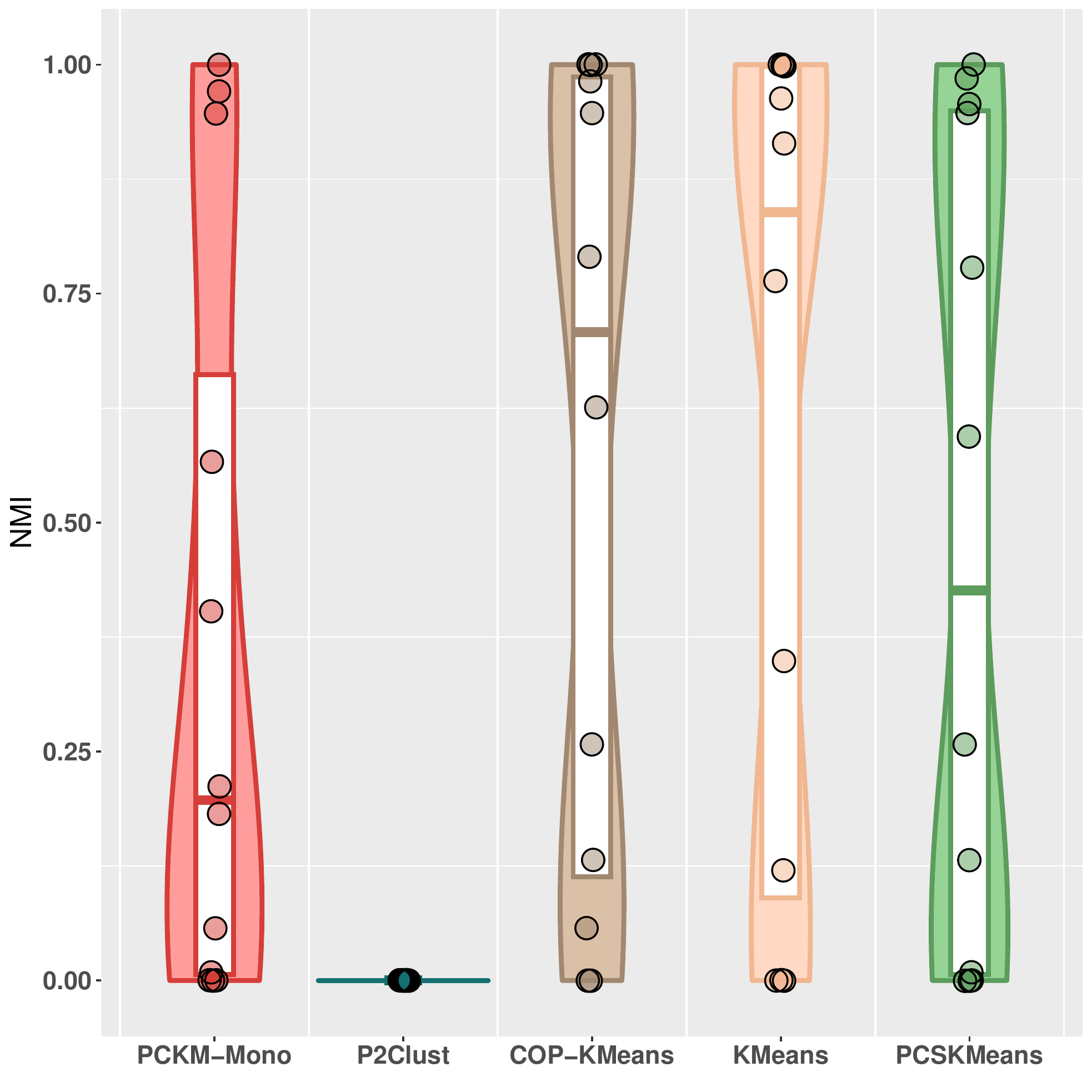}
		\label{fig:violin_nmi_15}}

    \vspace{\baselineskip}
  
	\subfloat[]{\includegraphics[width=0.8\linewidth]{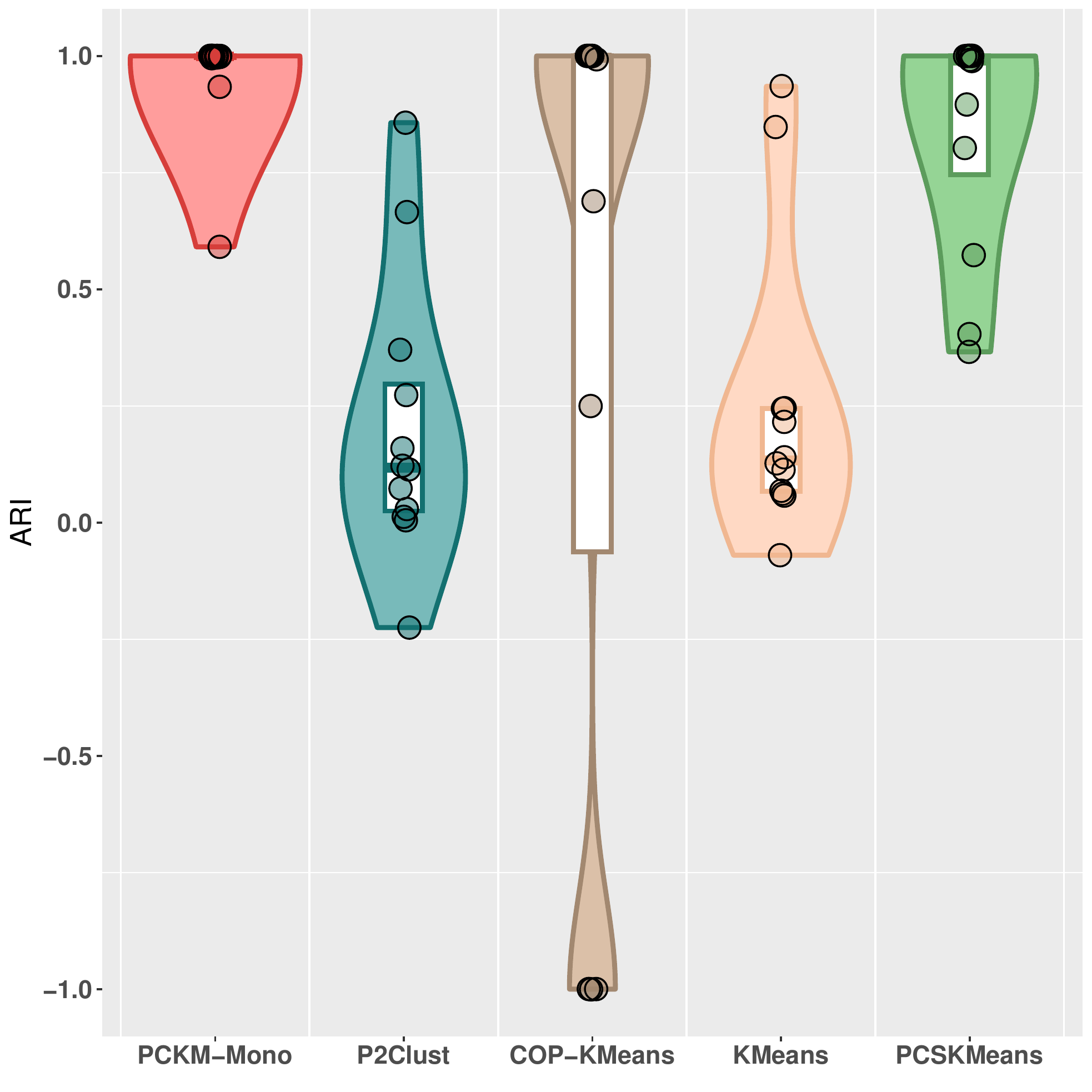}
		\label{fig:violin_ari_15}}

	\caption{Comparative violinplots for the results obtained with CS$_{15}$. Figures \ref{fig:violin_unsat_15}, \ref{fig:violin_nmi_15}, and \ref{fig:violin_ari_15} show the results obtained for the Unsat, NMI, and ARI measures, respectively.}
	\label{fig:ViolinPlots_CS15}
\end{figure}

\begin{figure}[!h]
	\centering
	 \subfloat[]{\includegraphics[width=0.4\linewidth]{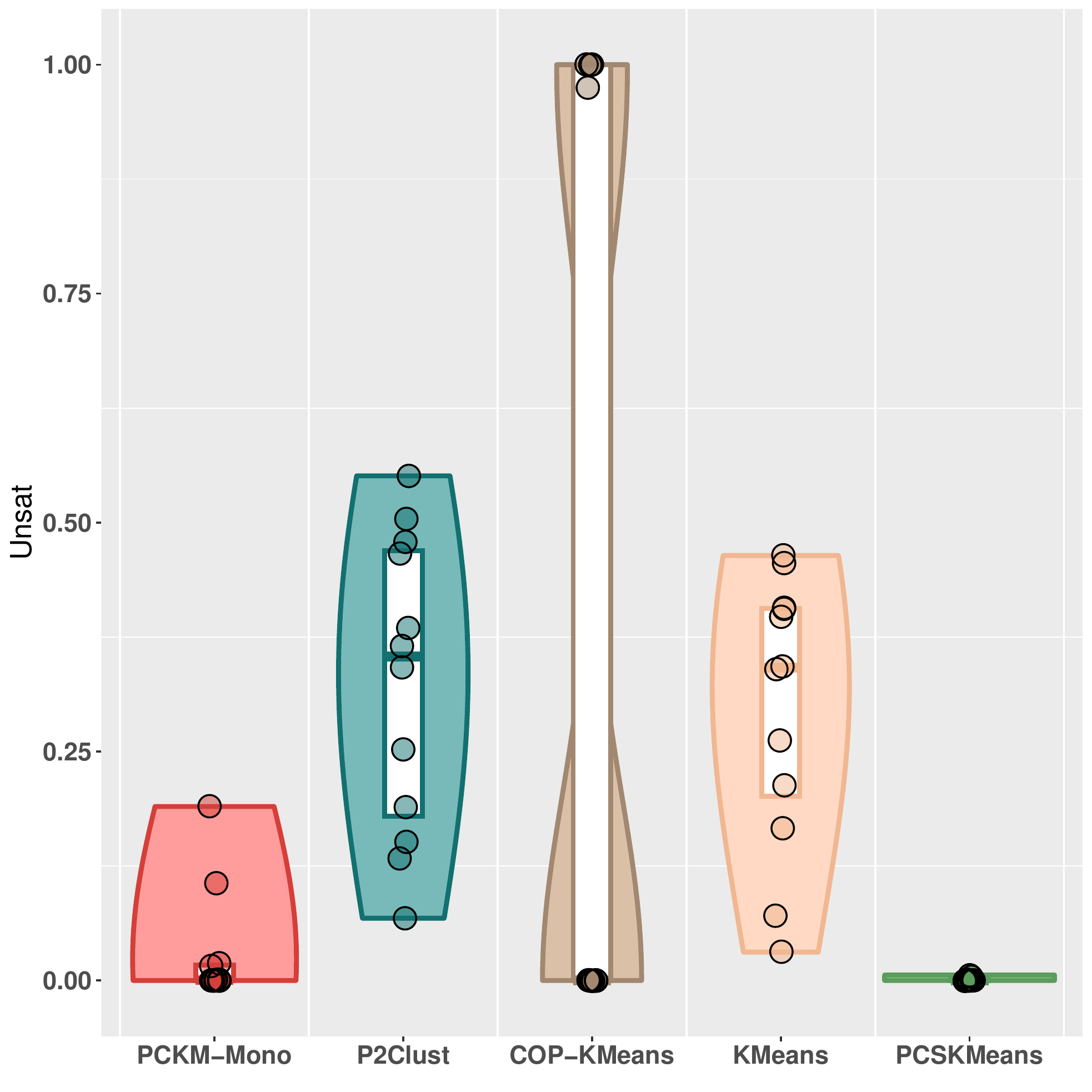}
		\label{fig:violin_unsat_20}}
	\subfloat[]{\includegraphics[width=0.4\linewidth]{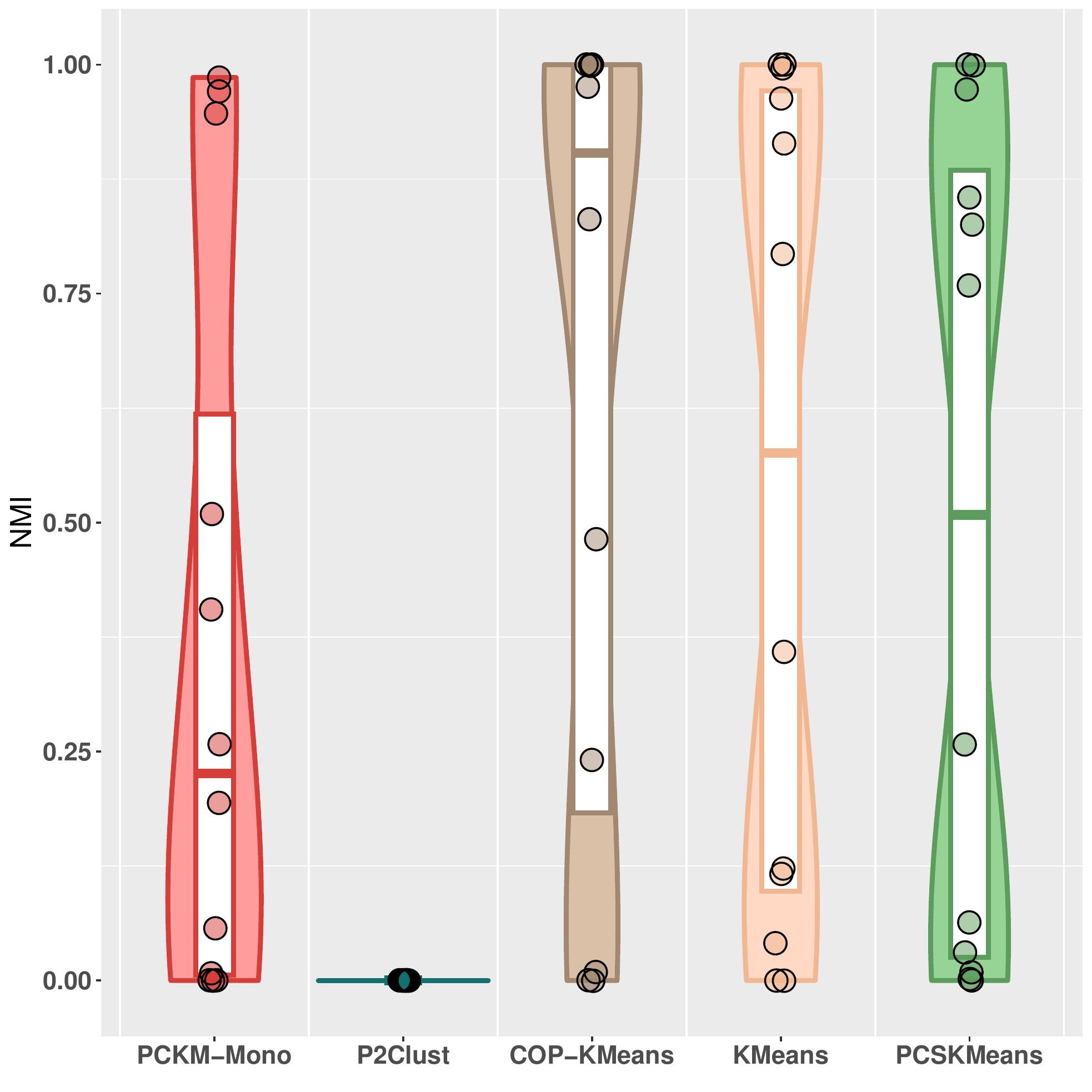}
		\label{fig:violin_nmi_20}}

    \vspace{\baselineskip}
  
	\subfloat[]{\includegraphics[width=0.8\linewidth]{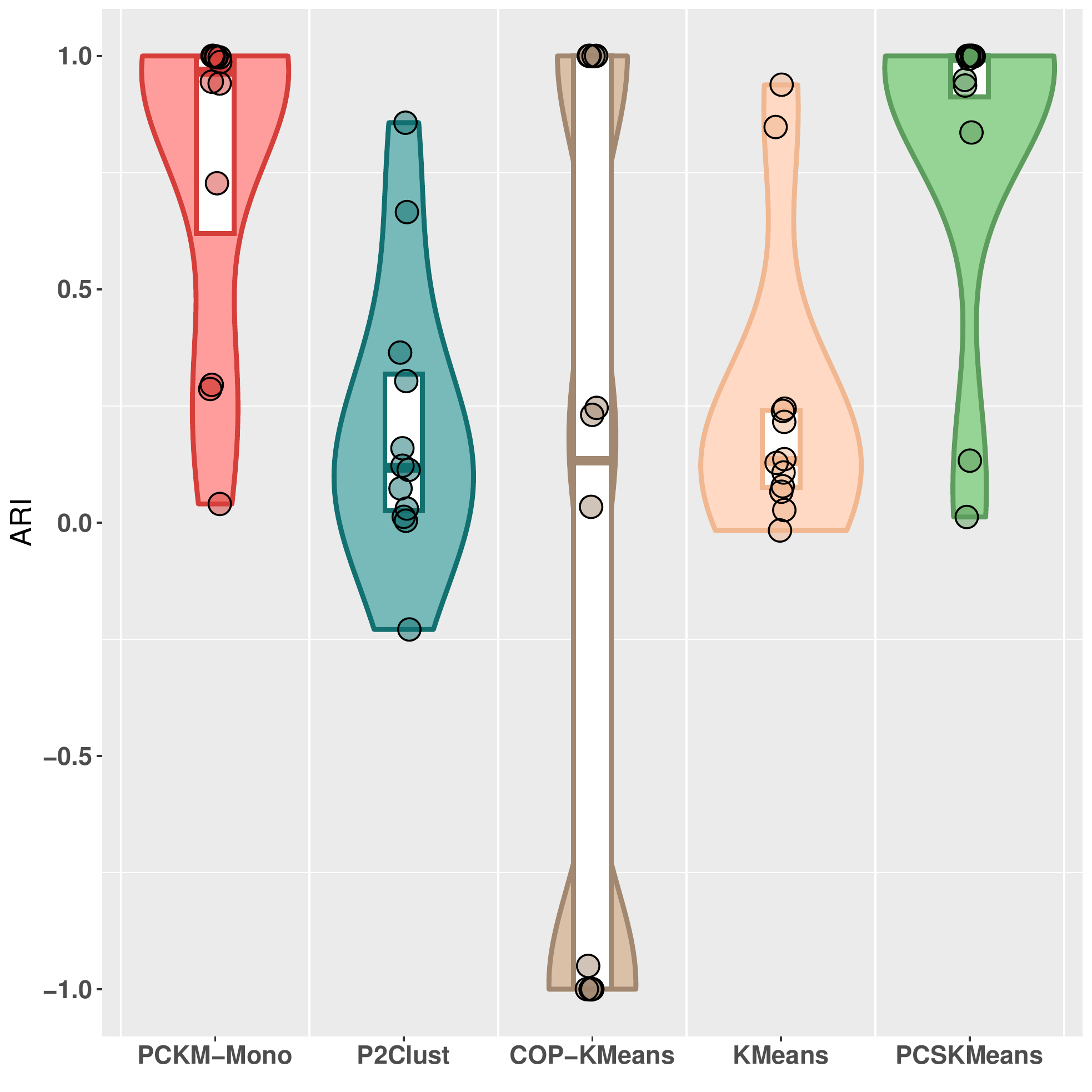}
		\label{fig:violin_ari_20}}

	\caption{Comparative violinplots for the results obtained with CS$_{20}$. Figures \ref{fig:violin_unsat_20}, \ref{fig:violin_nmi_20}, and \ref{fig:violin_ari_20} show the results obtained for the Unsat, NMI, and ARI measures, respectively.}
	\label{fig:ViolinPlots_CS20}
\end{figure}

Figures \ref{fig:ViolinPlots_CS10}, \ref{fig:ViolinPlots_CS15} and \ref{fig:ViolinPlots_CS20} are used to compare average results for all methods, and we refer to them as violinplots. They allow for a quick view of the distribution of results achieved by each method, as they contain a boxplot in addition to the outer violinplot.

By examining the results, it seems obvious that the proposed algorithm, PCKM-Mono, is able to find a balance between constraint satisfaction and the monotonicity of the output partition. Clearly P2Clust, which is a purely monotonic algorithm, always produces the best results with respect to NMI, as shown in Figures \ref{fig:violin_nmi_10}, \ref{fig:violin_nmi_15}, and \ref{fig:violin_nmi_20}. Similarly, Figures \ref{fig:violin_unsat_10}, \ref{fig:violin_unsat_15}, and \ref{fig:violin_unsat_20} show how purely constrained clustering algorithms (COP-Kmeans and PCSKMeans) produce the best results with respect to Unsat. However, PCKM-Mono is able to produce the best average ARI results (see Figures \ref{fig:violin_ari_10}, \ref{fig:violin_ari_15}, and \ref{fig:violin_ari_20}), while also achieving better NMI results than purely constrained clustering algorithms, and better Unsat results than purely monotonic clustering algorithms. This is indicative of the viability of the combination of pairwise and monotonic constraints to solve benchmark problems in both areas; moreover, it provides evidence in favor of the proposed EM optimization scheme, which is simple but can be, nonetheless, suitable for this task.

Some of the particular numerical results are worth noting, for example: the COP-Kmeans algorithm achieves near-optimum results for the CS$_{10}$ constraint set. The reason for this being that, the lower the number of constraints, the easier it is for the algorithm to find a feasible partition, which is usually a very accurate partition in the case of COP-Kmeans. With regard to the results obtained by PCKM-Mono for Unsat and NMI, both are observed to be stable with the increasing amount of constraint based information, while the ARI is observed to scale with it (although not in a consistent manner). Please note that, the results obtained by Kmeans and P2Clust are practically identical, independent of the constraint set, which is  a virtually average result, as they are not affected at all by constraints.

\begin{table}[!h]
	\centering
	\setlength{\tabcolsep}{5pt}
	\renewcommand{\arraystretch}{1.8}
	\caption{Results obtained by the five compared methods for the CS$_{10}$ constraint set.}
	\resizebox{\textwidth}{!}{
		\begin{tabular}{l ccccc c ccccc c ccccc}
		\hline
		
		Dataset & \multicolumn{5}{c}{ARI ($\uparrow$)} &&
		\multicolumn{5}{c}{NMI ($\downarrow$)} && \multicolumn{5}{c}{Unsat ($\downarrow$)} \\
		\cline{2-6} \cline{8-12} \cline{14-18}
		& PCKM-Mono & P2Clust & COP-KMeans & KMeans & PCSKMeans && PCKM-Mono & P2Clust & COP-KMeans & KMeans & PCSKMeans && PCKM-Mono & P2Clust & COP-KMeans & KMeans & PCSKMeans \\
		\hline
		Artiset & \textbf{1.000} & 0.366 & - & 0.241 & 0.995 && 0.020 & \textbf{0.000} & - & 0.810 & 0.189 && \textbf{0.000} & 0.136 & - & 0.160 & \textbf{0.000} \\
        Balance & 0.016 & 0.005 & \textbf{1.000} & 0.146 & \textbf{1.000} && 0.610 & \textbf{0.000} & 0.767 & 0.914 & 0.631 && 0.053 & 0.477 & \textbf{0.000} & 0.406 & \textbf{0.000} \\
        Bostonhousing4Cl & 0.123 & 0.122 & \textbf{1.000} & 0.124 & 0.657 && \textbf{0.000} & \textbf{0.000} & \textbf{0.000} & \textbf{0.000} & \textbf{0.000} && 0.091 & 0.332 & \textbf{0.000} & 0.356 & 0.005 \\
        Car & 0.825 & 0.036 & \textbf{1.000} & 0.112 & 0.993 && 0.058 & \textbf{0.000} & 0.057 & 0.128 & 0.164 && 0.008 & 0.500 & \textbf{0.000} & 0.458 & \textbf{0.000} \\
        ERA & 0.996 & 0.013 & - & -0.045 & \textbf{0.997} && 1.000 & \textbf{0.000} & - & 1.000 & 1.000 && \textbf{0.000} & 0.256 & - & 0.283 & \textbf{0.000} \\
        ESL & \textbf{0.979} & 0.281 & 0.975 & 0.241 & 0.974 && 0.584 & \textbf{0.000} & 0.653 & 1.000 & 0.584 && 0.001 & 0.210 & \textbf{0.000} & 0.206 & \textbf{0.000} \\
        LEV & \textbf{1.000} & -0.224 & \textbf{1.000} & 0.071 & 0.999 && 0.989 & \textbf{0.000} & 0.971 & 1.000 & 0.971 && \textbf{0.000} & 0.558 & \textbf{0.000} & 0.345 & \textbf{0.000} \\
        MachineCPU & 0.156 & 0.159 & \textbf{0.987} & 0.224 & 0.196 && 0.143 & \textbf{0.000} & 0.258 & 0.364 & 0.258 && 0.034 & 0.385 & \textbf{0.000} & 0.339 & 0.011 \\
        Qualitative Bankruptcy & \textbf{1.000} & 0.665 & -0.300 & 0.934 & \textbf{1.000} && \textbf{0.000} & \textbf{0.000} & 0.650 & \textbf{0.000} & \textbf{0.000} && \textbf{0.000} & 0.143 & 0.650 & 0.035 & \textbf{0.000} \\
        SWD & 0.217 & 0.111 & \textbf{1.000} & 0.066 & 0.955 && 0.947 & \textbf{0.000} & 0.947 & 0.973 & 0.933 && 0.093 & 0.380 & \textbf{0.000} & 0.390 & 0.001 \\
        Windsor Housing & 0.984 & 0.073 & 0.994 & 0.064 & \textbf{1.000} && \textbf{0.000} & \textbf{0.000} & \textbf{0.000} & \textbf{0.000} & \textbf{0.000} && 0.001 & 0.452 & \textbf{0.000} & 0.452 & \textbf{0.000} \\
        Wisconsin & \textbf{1.000} & 0.857 & \textbf{1.000} & 0.849 & \textbf{1.000} && 0.009 & \textbf{0.000} & 0.009 & 0.764 & 0.009 && \textbf{0.000} & 0.070 & \textbf{0.000} & 0.074 & \textbf{0.000} \\
        Mean & 0.691 & 0.205 & 0.555 & 0.252 & \textbf{0.897} && 0.363 & \textbf{0.000} & 0.526 & 0.579 & 0.395 && 0.023 & 0.325 & 0.221 & 0.292 & \textbf{0.001} \\

        \hline
    	\end{tabular}}
	\label{tab:res10}
\end{table}

\begin{table}[!h]
	\centering
	\setlength{\tabcolsep}{5pt}
	\renewcommand{\arraystretch}{1.8}
	\caption{Results obtained by the five compared methods for the CS$_{15}$ constraint set.}
	\resizebox{\textwidth}{!}{
		\begin{tabular}{l ccccc c ccccc c ccccc}
		\hline
		Dataset & \multicolumn{5}{c}{ARI ($\uparrow$)} && \multicolumn{5}{c}{NMI ($\downarrow$)} && \multicolumn{5}{c}{Unsat ($\downarrow$)} \\
		\cline{2-6} \cline{8-12} \cline{14-18}
		& PCKM-Mono & P2Clust & COP-KMeans & KMeans & PCSKMeans && PCKM-Mono & P2Clust & COP-KMeans & KMeans & PCSKMeans && PCKM-Mono & P2Clust & COP-KMeans & KMeans & PCSKMeans \\
		\hline
		Artiset & 0.591 & 0.370 & \textbf{1.000} & 0.244 & 0.404 && 0.182 & \textbf{0.000} & \textbf{0.000} & 0.999 & 0.957 && 0.002 & 0.134 & \textbf{0.000} & 0.157 & 0.001 \\
        Balance & \textbf{1.000} & 0.005 & \textbf{1.000} & 0.140 & \textbf{1.000} && 0.566 & \textbf{0.000} & 0.790 & 0.914 & 0.778 && \textbf{0.000} & 0.482 & \textbf{0.000} & 0.392 & \textbf{0.000} \\
        Bostonhousing4Cl & 0.999 & 0.122 & \textbf{1.000} & 0.127 & 0.989 && \textbf{0.000} & \textbf{0.000} & \textbf{0.000} & \textbf{0.000} & \textbf{0.000} && \textbf{0.000} & 0.342 & \textbf{0.000} & 0.349 & \textbf{0.000} \\
        Car & 0.999 & 0.029 & \textbf{1.000} & 0.113 & \textbf{1.000} && 0.057 & \textbf{0.000} & 0.057 & 0.120 & 0.131 && \textbf{0.000} & 0.501 & \textbf{0.000} & 0.461 & \textbf{0.000} \\
        ERA & \textbf{0.998} & 0.012 & - & -0.070 & 0.573 && 1.000 & \textbf{0.000} & - & 1.000 & 1.000 && \textbf{0.000} & 0.252 & - & 0.307 & \textbf{0.000} \\
        ESL & \textbf{0.995} & 0.273 & 0.993 & 0.245 & 0.366 && 0.403 & \textbf{0.000} & 0.626 & 0.998 & 0.594 && \textbf{0.000} & 0.209 & \textbf{0.000} & 0.211 & 0.001 \\
        LEV & \textbf{0.934} & -0.225 & 0.250 & 0.062 & 0.896 && 0.971 & \textbf{0.000} & 0.982 & 1.000 & 0.985 && 0.003 & 0.555 & 0.375 & 0.340 & \textbf{0.000} \\
        MachineCPU & \textbf{1.000} & 0.159 & \textbf{1.000} & 0.216 & 0.803 && 0.212 & \textbf{0.000} & 0.258 & 0.349 & 0.258 && \textbf{0.000} & 0.344 & \textbf{0.000} & 0.377 & \textbf{0.000} \\
        Qualitative Bankruptcy & \textbf{1.000} & 0.665 & 0.689 & 0.935 & \textbf{1.000} && \textbf{0.000} & \textbf{0.000} & 0.132 & \textbf{0.000} & \textbf{0.000} && \textbf{0.000} & 0.173 & 0.125 & 0.038 & \textbf{0.000} \\
        SWD & 0.997 & 0.114 & \textbf{1.000} & 0.068 & \textbf{1.000} && 0.947 & \textbf{0.000} & 0.947 & 0.963 & 0.947 && 0.001 & 0.381 & \textbf{0.000} & 0.394 & \textbf{0.000} \\
        Windsor Housing & \textbf{1.000} & 0.073 & - & 0.058 & 0.992 && \textbf{0.000} & \textbf{0.000} & - & \textbf{0.000} & \textbf{0.000} && \textbf{0.000} & 0.464 & - & 0.451 & \textbf{0.000} \\
        Wisconsin & \textbf{1.000} & 0.857 & - & 0.848 & \textbf{1.000} && 0.009 & \textbf{0.000} & - & 0.764 & 0.009 && \textbf{0.000} & 0.069 & - & 0.075 & \textbf{0.000} \\
        Mean & \textbf{0.960} & 0.205 & 0.411 & 0.249 & 0.835 && 0.362 & \textbf{0.000} & 0.566 & 0.592 & 0.472 && 0.001 & 0.326 & 0.292 & 0.296 & \textbf{0.000} \\

        \hline
    	\end{tabular}}
	\label{tab:res15}
\end{table}

\begin{table}[!h]
	\centering
	\setlength{\tabcolsep}{5pt}
	\renewcommand{\arraystretch}{1.8}
	\caption{Results obtained by the five compared methods for the CS$_{20}$ constraint set.}
	\resizebox{\textwidth}{!}{
		\begin{tabular}{l ccccc c ccccc c ccccc}
		\hline
		Dataset & \multicolumn{5}{c}{ARI ($\uparrow$)} && \multicolumn{5}{c}{NMI ($\downarrow$)} && \multicolumn{5}{c}{Unsat ($\downarrow$)} \\
		\cline{2-6} \cline{8-12} \cline{14-18}
		& PCKM-Mono & P2Clust & COP-KMeans & KMeans & PCSKMeans && PCKM-Mono & P2Clust & COP-KMeans & KMeans & PCSKMeans && PCKM-Mono & P2Clust & COP-KMeans & KMeans & PCSKMeans \\
		\hline
		Artiset & 0.941 & 0.364 & -0.950 & 0.239 & \textbf{1.000} && 0.194 & \textbf{0.000} & 0.976 & 0.793 & 0.855 && 0.001 & 0.133 & 0.975 & 0.166 & \textbf{0.000} \\
        Balance & \textbf{1.000} & 0.004 & \textbf{1.000} & 0.136 & \textbf{1.000} && 0.509 & \textbf{0.000} & 0.831 & 0.914 & 0.825 && \textbf{0.000} & 0.479 & \textbf{0.000} & 0.407 & \textbf{0.000} \\
        Bostonhousing4Cl & \textbf{1.000} & 0.122 & - & 0.128 & \textbf{1.000} && \textbf{0.000} & \textbf{0.000} & - & \textbf{0.000} & \textbf{0.000} && \textbf{0.000} & 0.342 & - & 0.340 & \textbf{0.000} \\
        Car & 0.999 & 0.030 & - & 0.107 & \textbf{1.000} && 0.057 & \textbf{0.000} & - & 0.122 & 0.063 && \textbf{0.000} & 0.504 & - & 0.464 & \textbf{0.000} \\
        ERA & 0.040 & 0.012 & 0.034 & -0.017 & \textbf{0.999} && 0.986 & \textbf{0.000} & 0.999 & 1.000 & 0.999 && 0.019 & 0.252 & \textbf{0.000} & 0.262 & \textbf{0.000} \\
        ESL & 0.295 & 0.303 & 0.246 & 0.244 & \textbf{0.994} && 0.405 & \textbf{0.000} & 0.482 & 1.000 & 0.759 && \textbf{0.000} & 0.189 & \textbf{0.000} & 0.213 & \textbf{0.000} \\
        LEV & \textbf{0.997} & -0.229 & - & 0.078 & 0.012 && 0.971 & \textbf{0.000} & - & 0.996 & 0.973 && \textbf{0.000} & 0.551 & - & 0.343 & 0.001 \\
        MachineCPU & \textbf{0.987} & 0.159 & 0.231 & 0.216 & 0.949 && 0.258 & \textbf{0.000} & 0.241 & 0.359 & 0.258 && \textbf{0.000} & 0.365 & \textbf{0.000} & 0.406 & \textbf{0.000} \\
        Qualitative Bankruptcy & 0.727 & 0.665 & \textbf{1.000} & 0.938 & 0.937 && \textbf{0.000} & \textbf{0.000} & \textbf{0.000} & 0.116 & 0.031 && 0.106 & 0.151 & \textbf{0.000} & 0.031 & \textbf{0.000} \\
        SWD & 0.998 & 0.113 & - & 0.066 & \textbf{1.000} && 0.947 & \textbf{0.000} & - & 0.963 & 1.000 && 0.001 & 0.385 & - & 0.397 & \textbf{0.000} \\
        Windsor Housing & 0.286 & 0.073 & \textbf{1.000} & 0.027 & 0.133 && \textbf{0.000} & \textbf{0.000} & \textbf{0.000} & \textbf{0.000} & 0.002 && 0.190 & 0.466 & \textbf{0.000} & 0.456 & 0.006 \\
        Wisconsin & 0.945 & 0.857 & \textbf{1.000} & 0.848 & 0.836 && 0.008 & \textbf{0.000} & 0.009 & 0.041 & 0.009 && 0.016 & 0.068 & \textbf{0.000} & 0.071 & 0.004 \\
        Mean & 0.768 & 0.206 & -0.037 & 0.251 & \textbf{0.822} && 0.361 & \textbf{0.000} & 0.628 & 0.525 & 0.481 && 0.028 & 0.324 & 0.415 & 0.296 & \textbf{0.001} \\
        \hline
    	\end{tabular}}
	\label{tab:res20}
\end{table}

\clearpage

\section{Statistical Analysis of Results} \label{sec:statisticalAnalysis}

\begin{figure}[!h]
	\centering
	\subfloat[ ]{\includegraphics[width=0.25\linewidth]{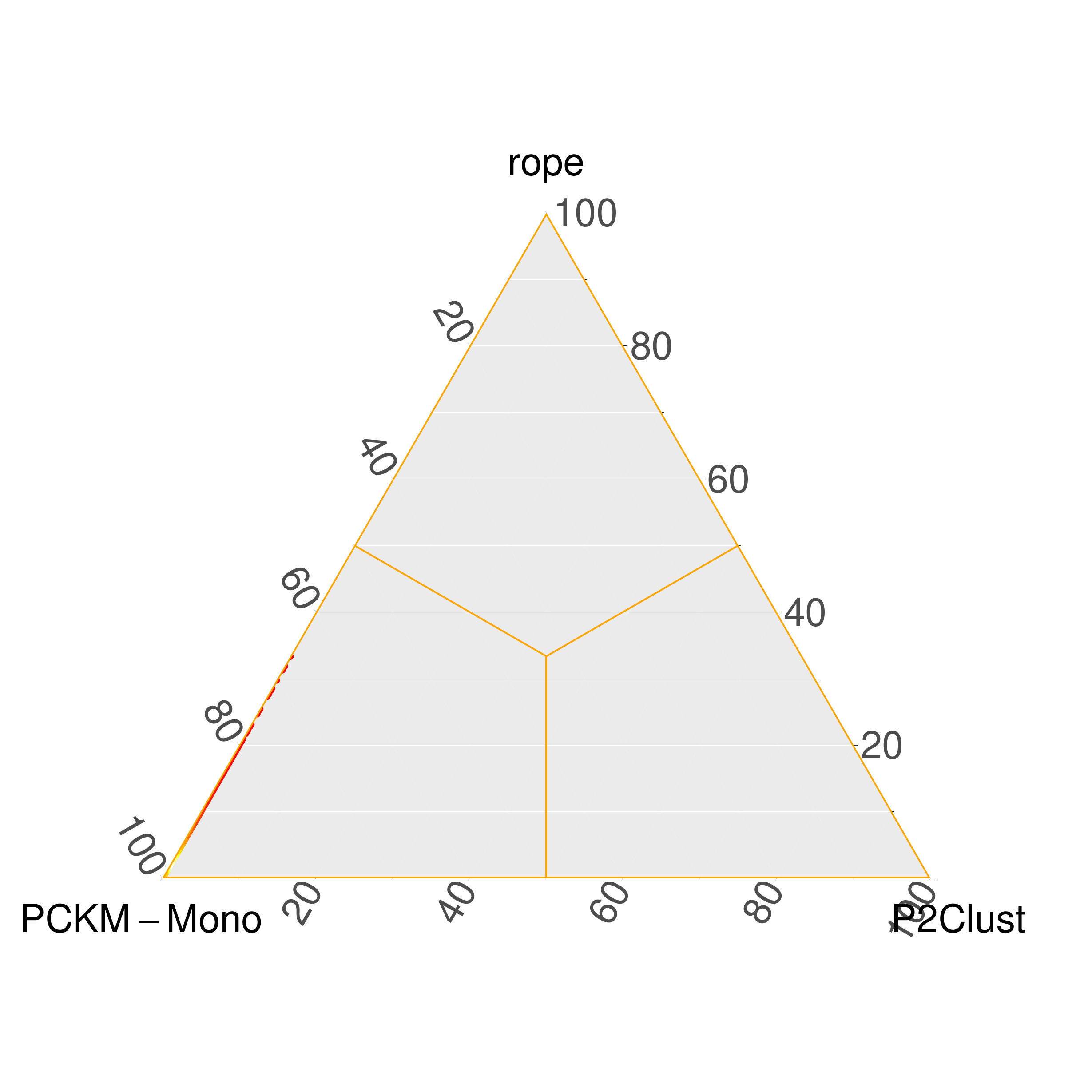}
		\label{fig:PCKMMonovsP2Clust_ARI}}
	\subfloat[ ]{\includegraphics[width=0.25\linewidth]{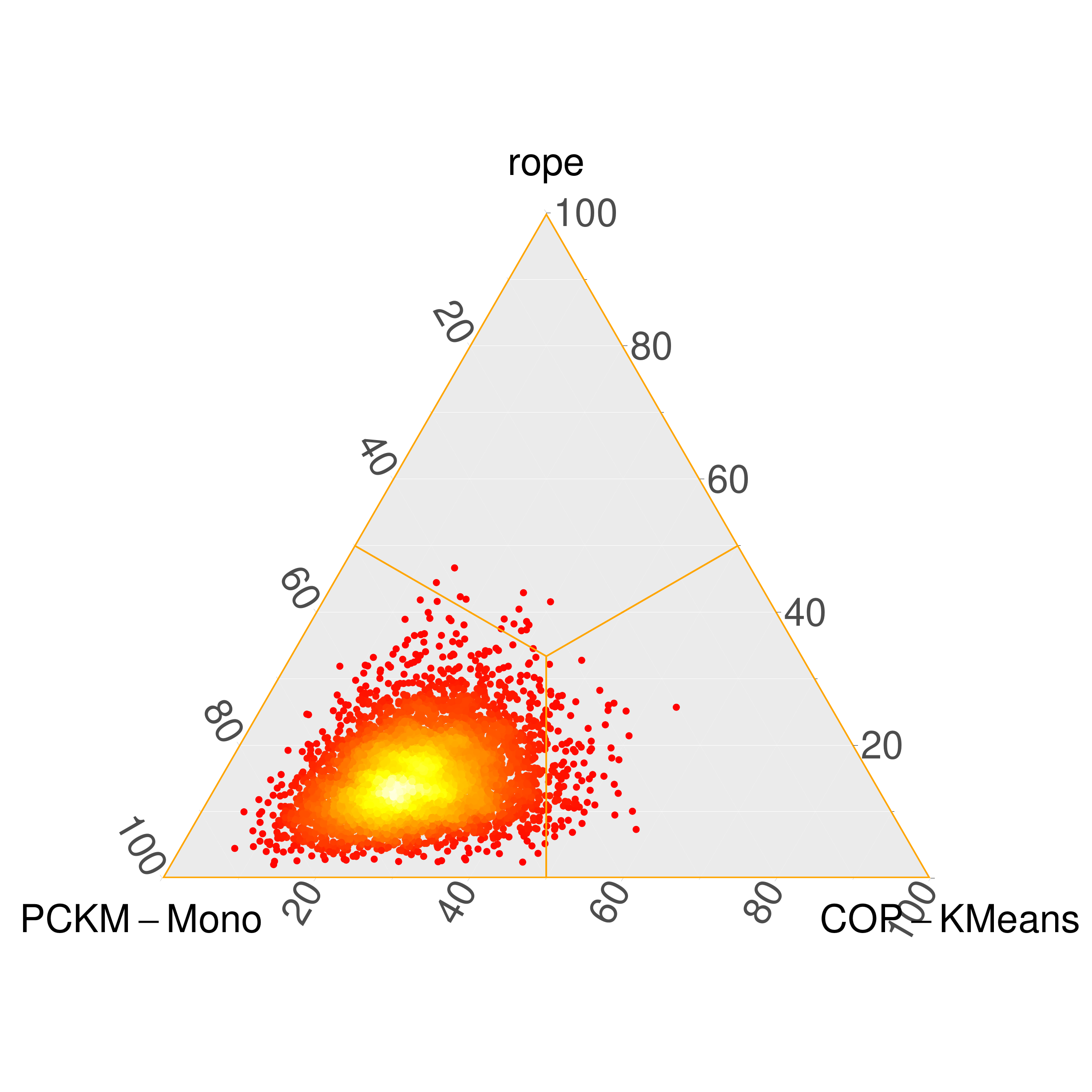}
		\label{fig:PCKMMonovsCOPKMeans_ARI}}
	\subfloat[ ]{\includegraphics[width=0.25\linewidth]{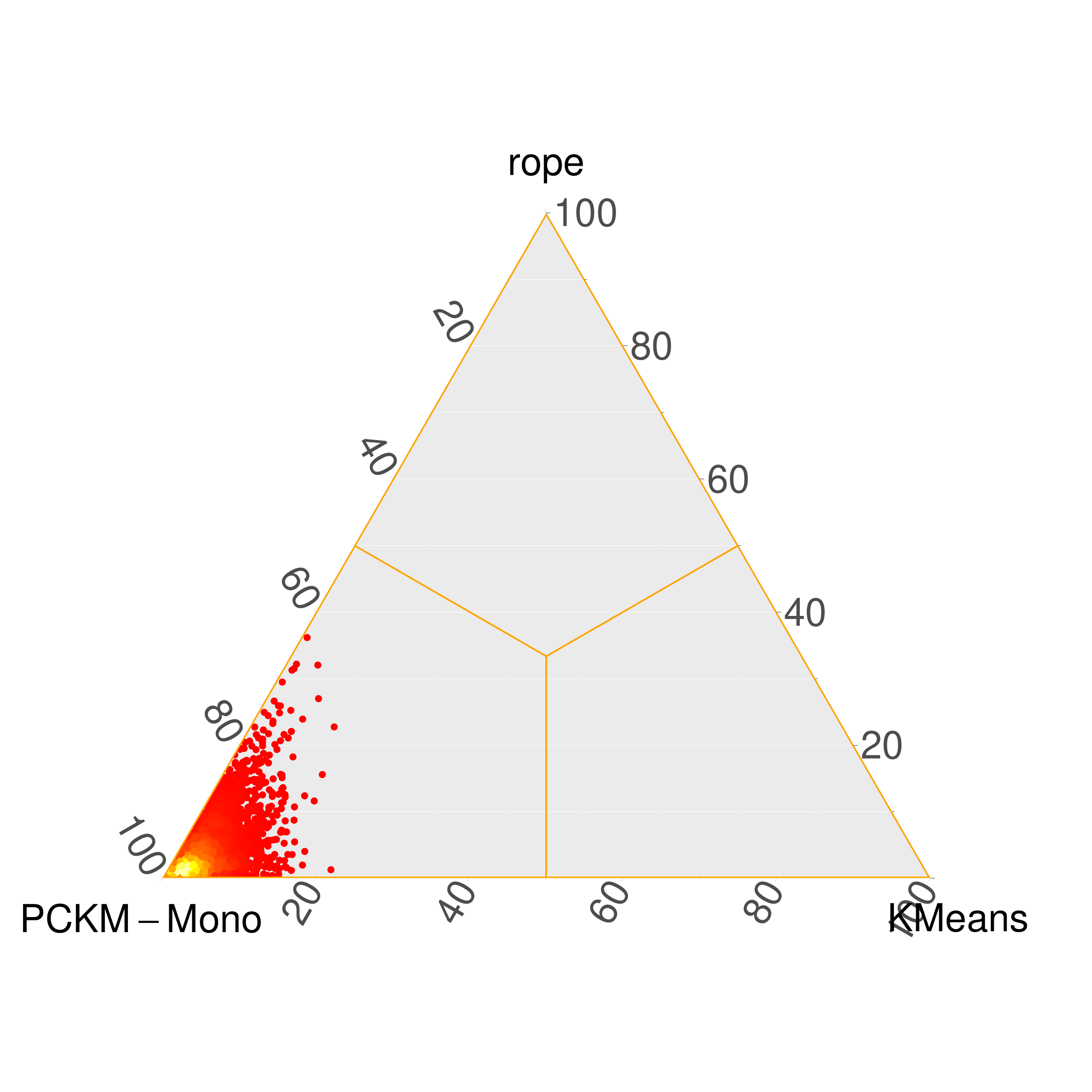}
		\label{fig:PCKMMonovsKMeans_ARI}}
	\subfloat[ ]{\includegraphics[width=0.25\linewidth]{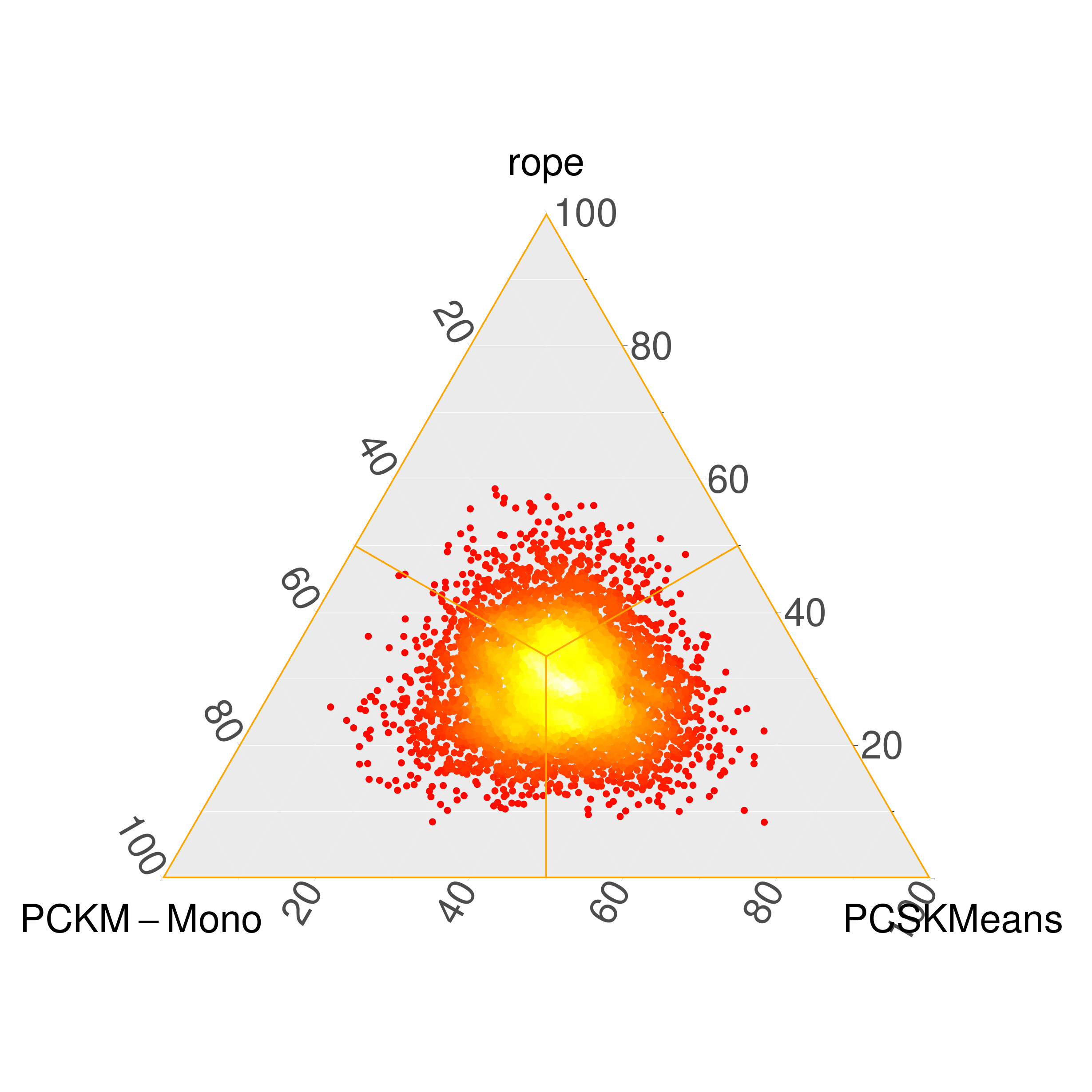}
		\label{fig:PCKMMonovsPCSKMeans_ARI}}
	\vspace{\baselineskip}
	\subfloat[ ]{\includegraphics[width=0.25\linewidth]{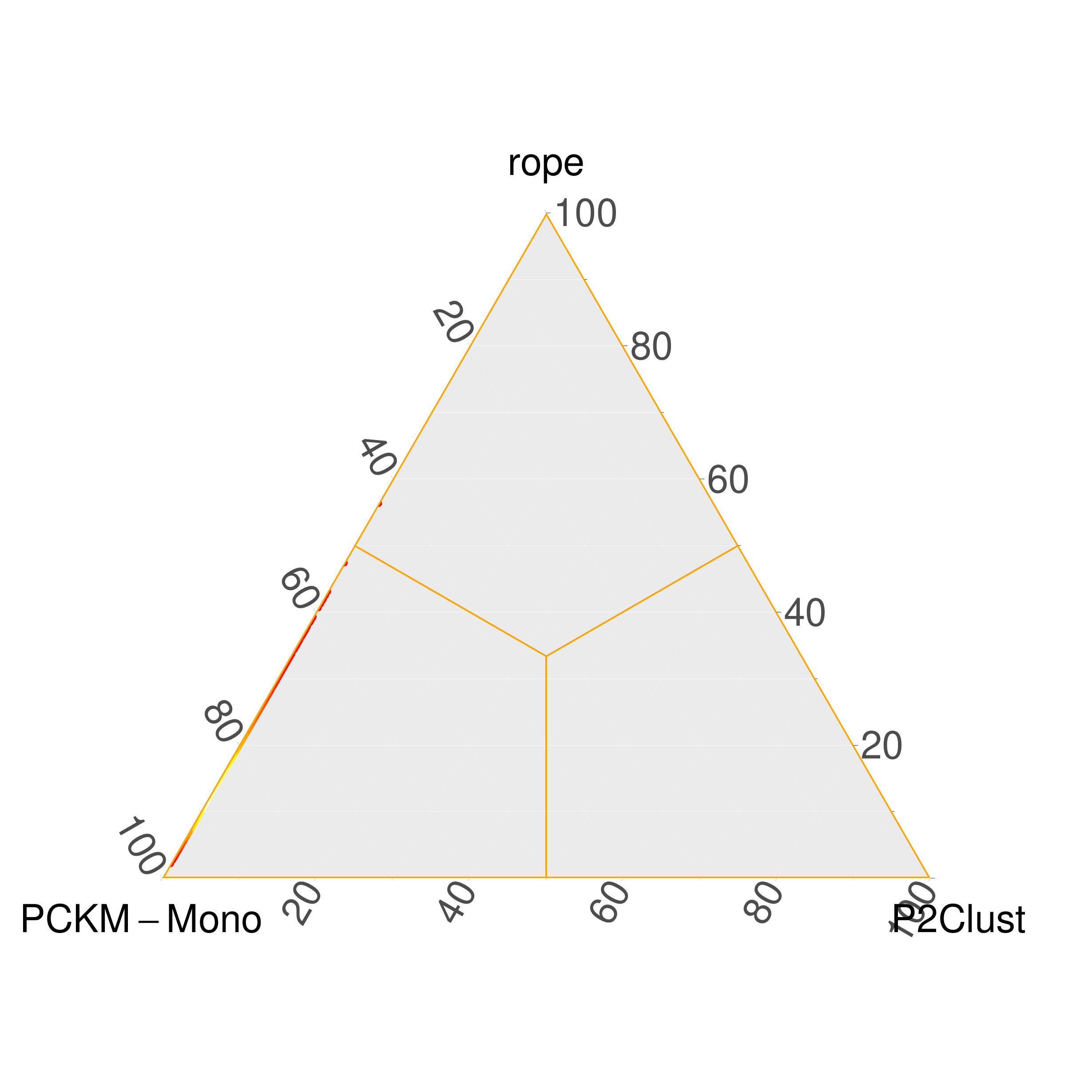}
		\label{fig:PCKMMonovsP2Clust_NMI}}
	\subfloat[ ]{\includegraphics[width=0.25\linewidth]{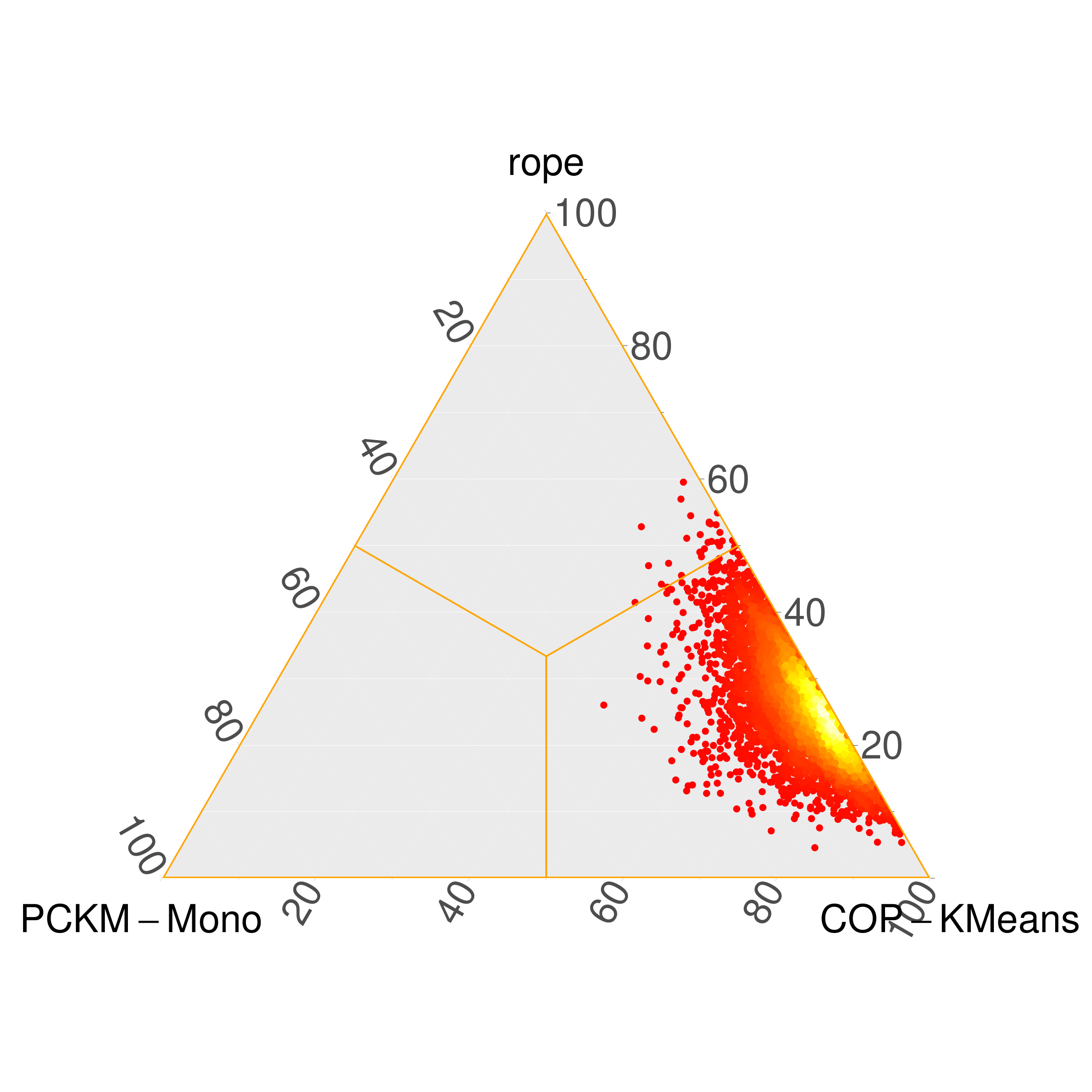}
		\label{fig:PCKMMonovsCOPKMeans_NMI}}
	\subfloat[ ]{\includegraphics[width=0.25\linewidth]{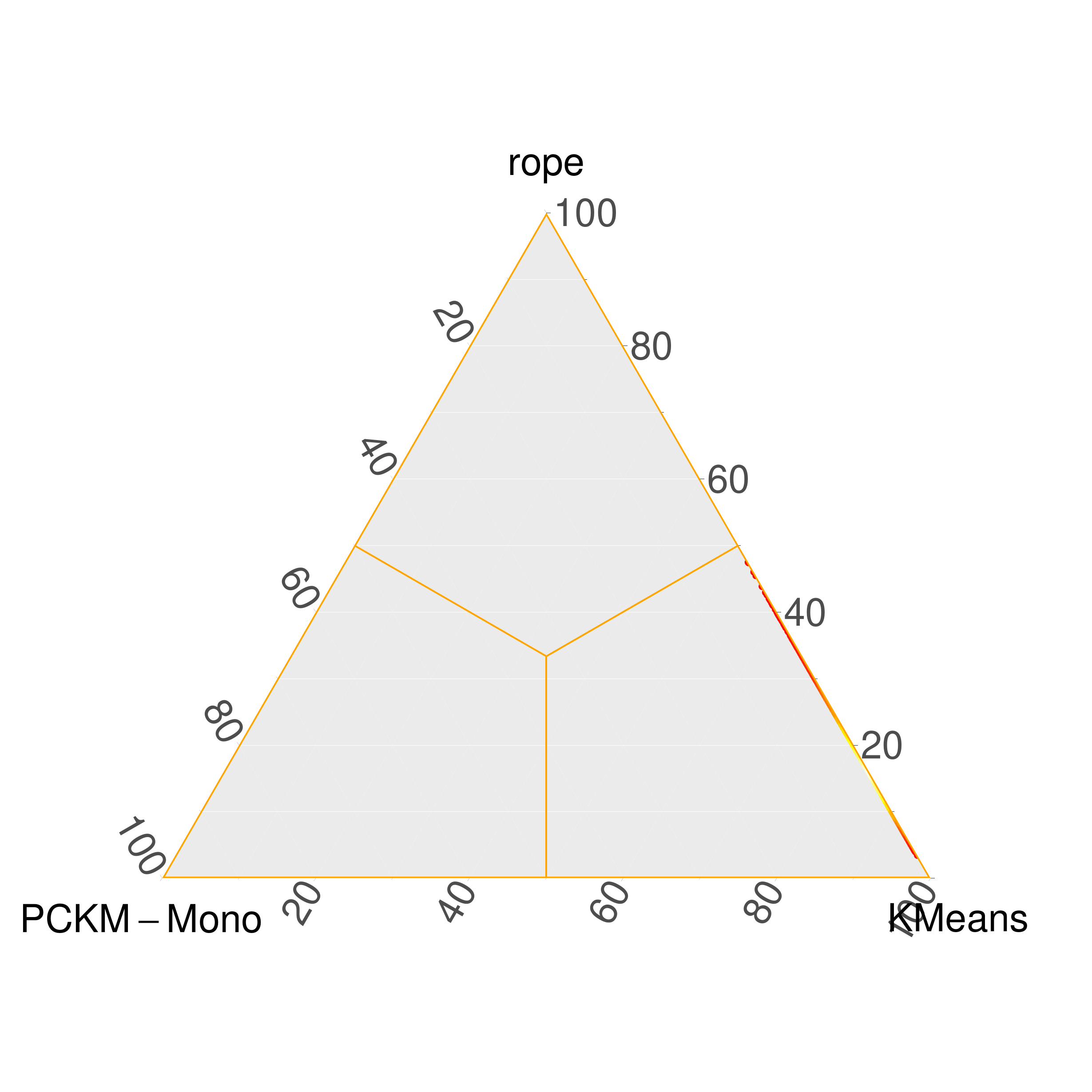}
		\label{fig:PCKMMonovsKMeans_NMI}}
	\subfloat[ ]{\includegraphics[width=0.25\linewidth]{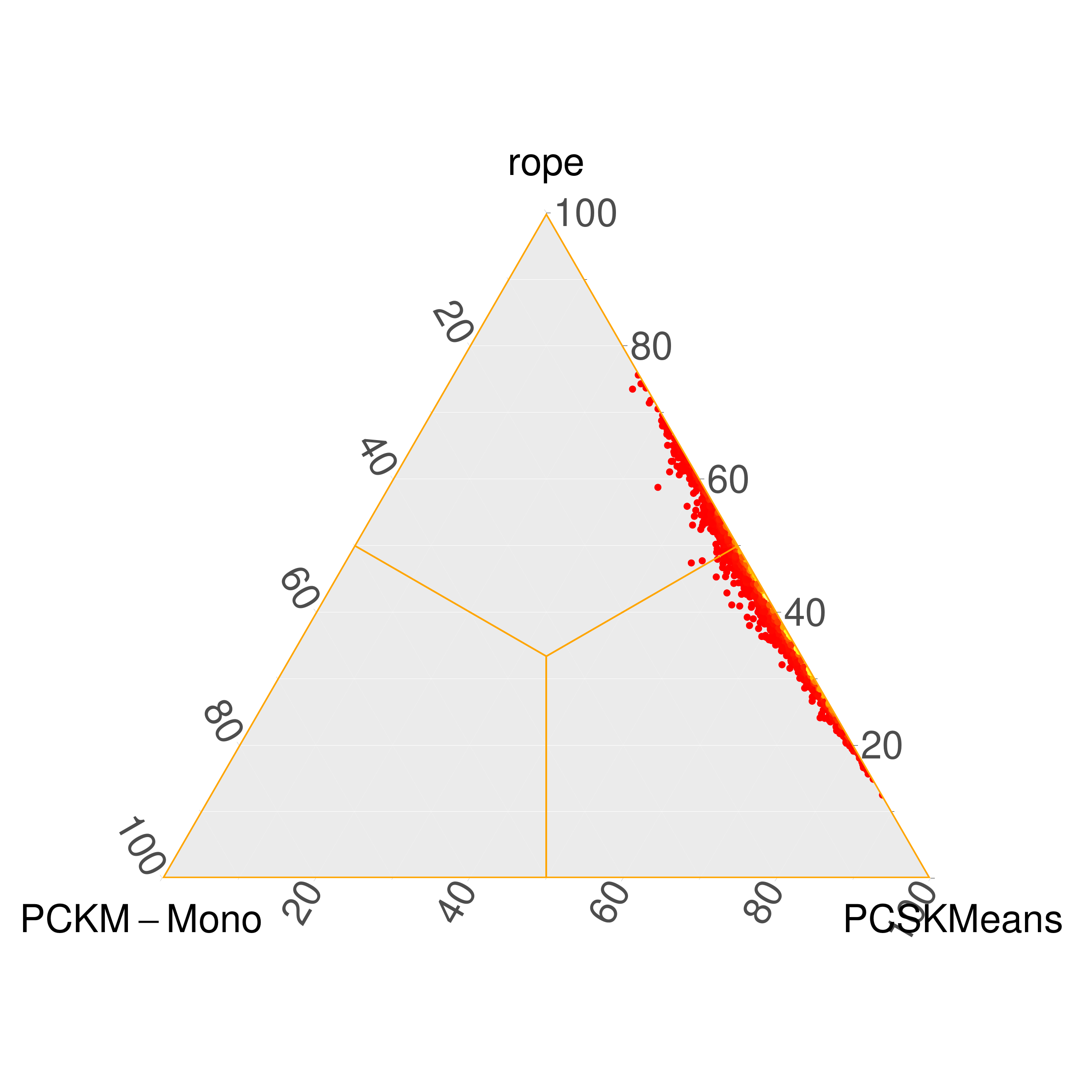}
		\label{fig:PCKMMonovsPCSKMeans_NMI}}
	\vspace{\baselineskip}
	\subfloat[ ]{\includegraphics[width=0.25\linewidth]{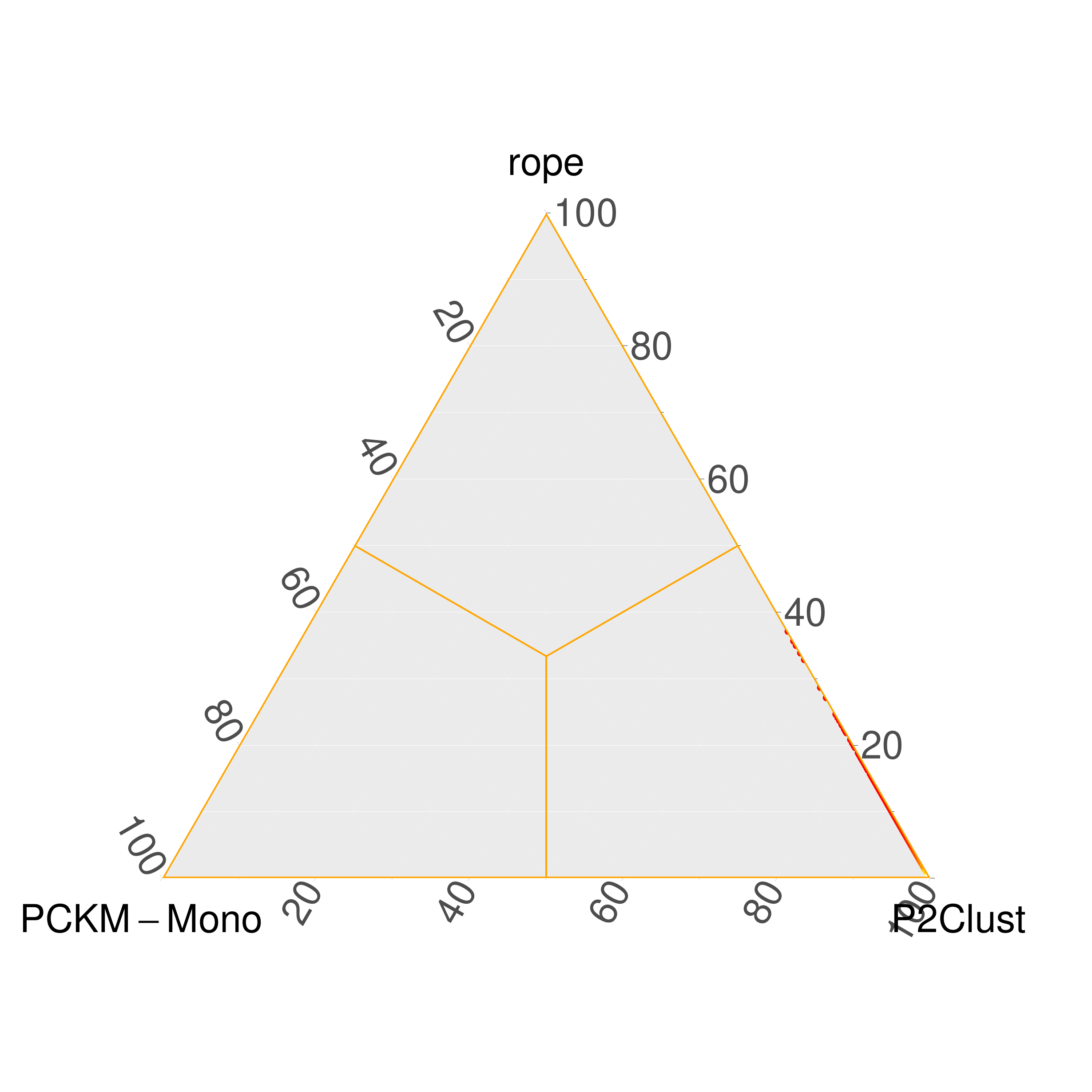}
		\label{fig:PCKMMonovsP2Clust_Unsat}}
	\subfloat[ ]{\includegraphics[width=0.25\linewidth]{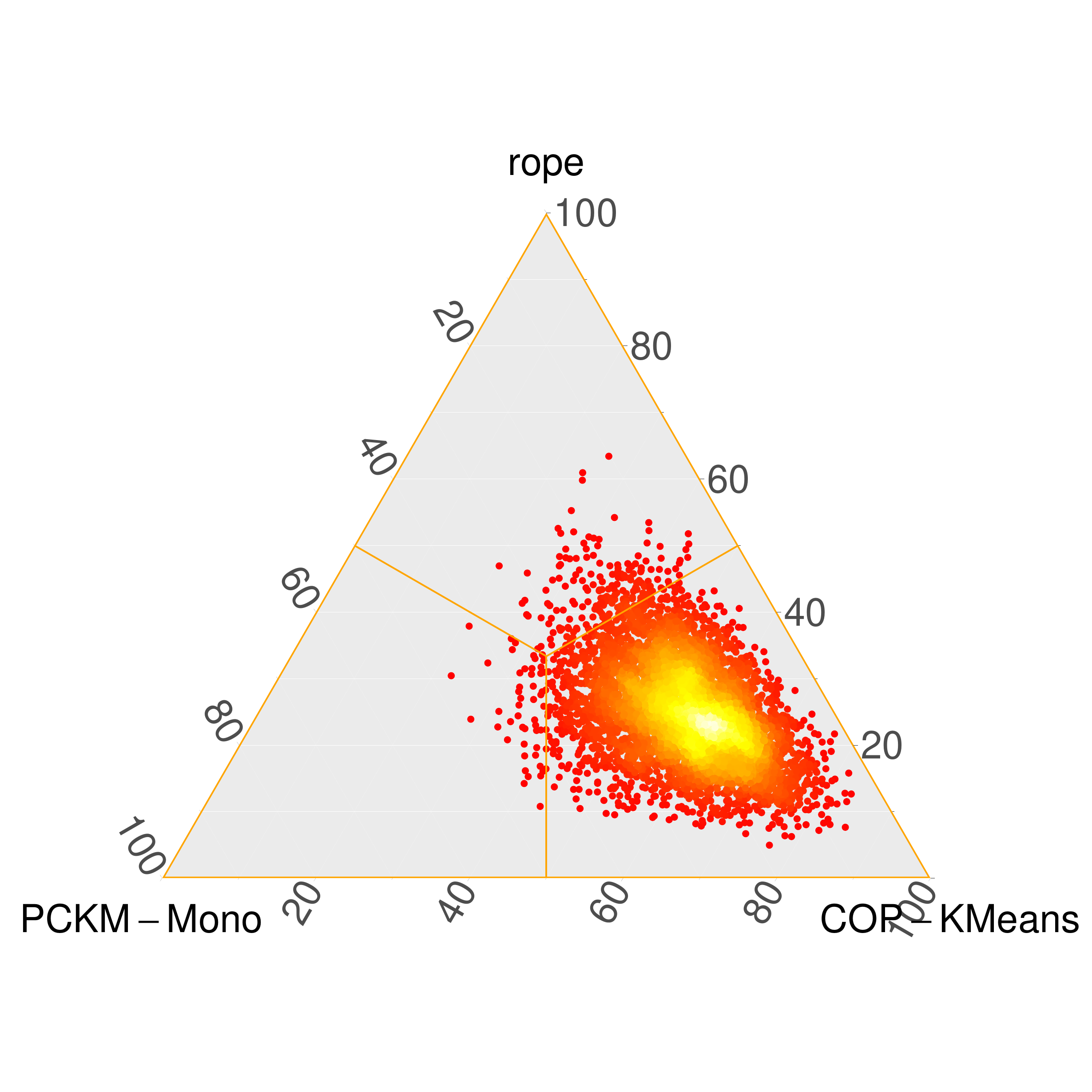}
		\label{fig:PCKMMonovsCOPKMeans_Unsat}}
	\subfloat[ ]{\includegraphics[width=0.25\linewidth]{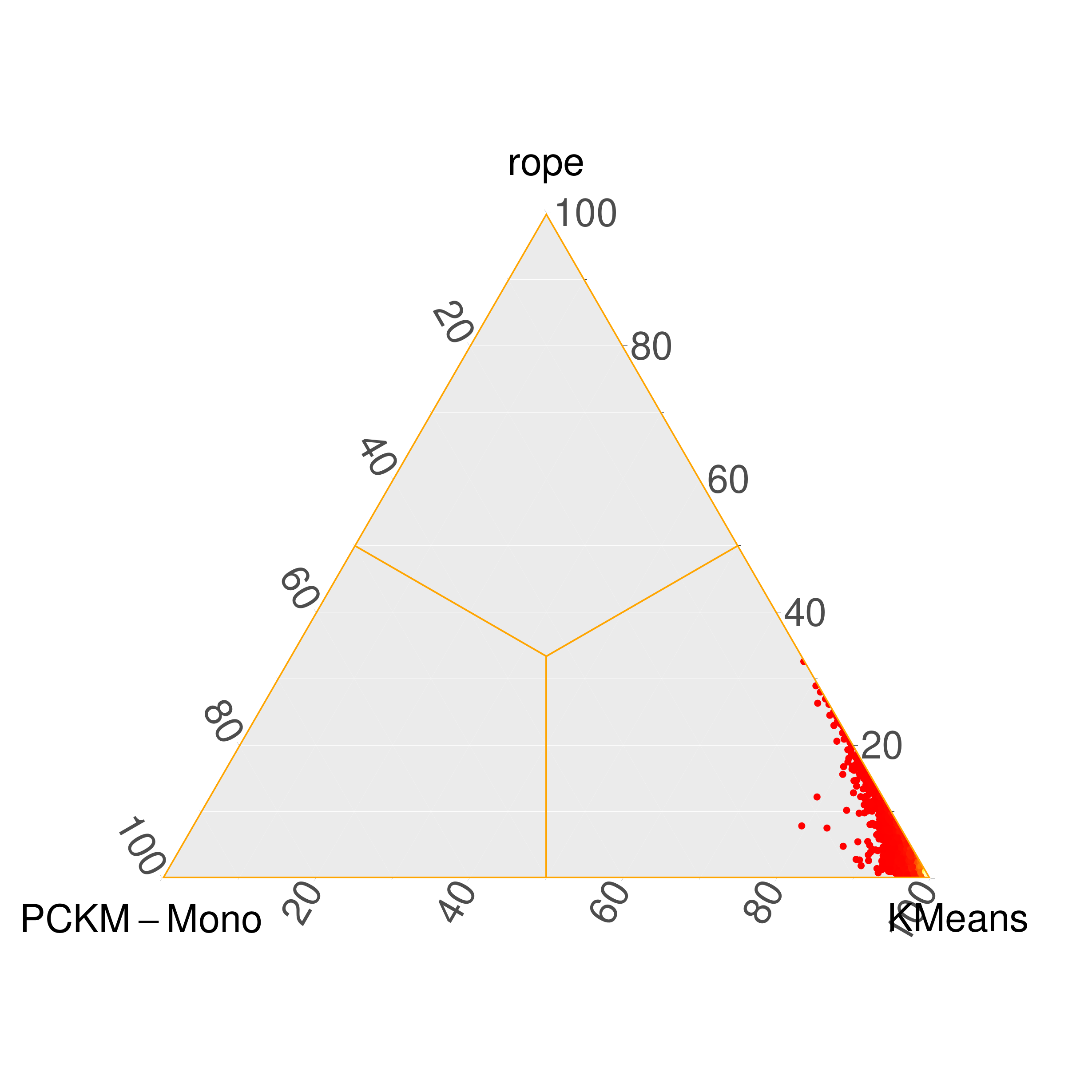}
		\label{fig:PCKMMonovsKMeans_Unsat}}
	\subfloat[ ]{\includegraphics[width=0.25\linewidth]{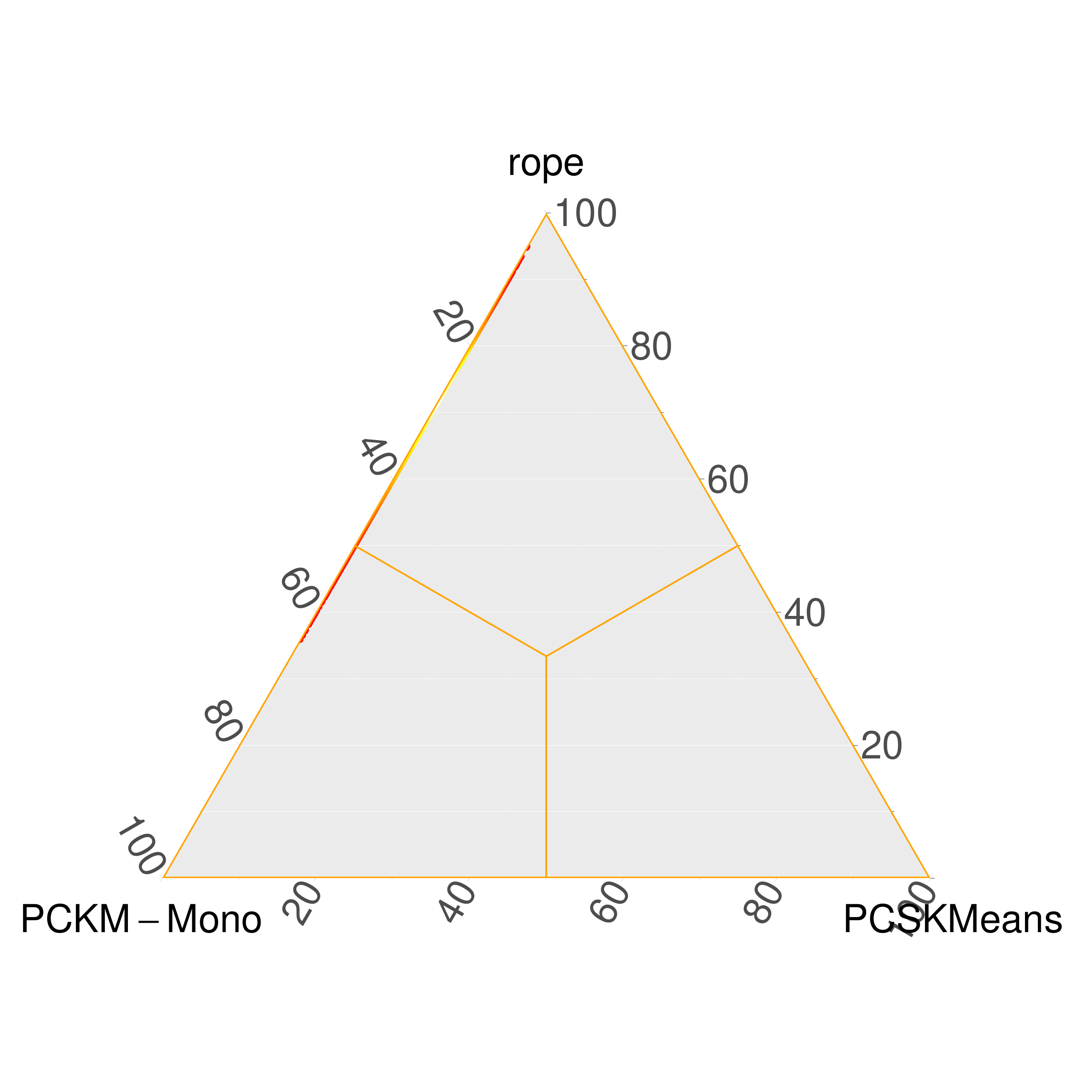}
		\label{fig:PCKMMonovsPCSKMeans_Unsat}}

	\caption{Heatmaps \ref{fig:PCKMMonovsP2Clust_ARI} to \ref{fig:PCKMMonovsPCSKMeans_Unsat} compare the proposed method PCKM-Mono to the other four compared methods: P2Clust, COP-Kmeans, Kmeans and PCSKMeans, respectively, from left to right in every row. The first row compares the results for the ARI measure, the second row does so for NMI and the third is for Unsat.}
	\label{fig:Heatmaps}
\end{figure}

In contrast with NHST, it is possible to create illustrative graphical representations of the results of the Bayesian sign test. To do so, the obtained distribution is sampled to obtain a set of triplets, which are interpreted as barycentric coordinates in an equilateral triangle, thus producing a cloud of points with varying density. This is known as a heatmap. Figure \ref{fig:Heatmaps} shows heatmaps which compare the proposed method PCKM-Mono with the rest of the benchmarked methods for the three measures obtained: ARI, NMI and Unsat. The region of practical equivalence is set to $rope = [-0.02, 0.02]$ for ARI, and to $rope = [-0.01, 0.01]$ for NMI and Unsat, following the guidelines in \cite{carrasco2020recent}. The results produced by PCKM-Mono are always taken as $B$ in \ref{eq:BST_triplet}, and $A$ represents the set of results obtained by the compared method. Please note that, as ARI is a measure to maximize, a cloud of points located in the region of the map corresponding to MPCK-Means would indicate statistically significant differences between the two methods in favor of MPCK-Means. The opposite situation is found for NMI and Unsat.

All heatmaps reinforce the conclusions obtained in the Experimental Results Section \ref{sec:ExpResults}. It is clear that PCKM-Mono represents a statistically significant improvement over all compared method with respect to ARI, except for the comparison against PCSKMeans, which is the most debated one with a slight advantage for PCSKMeans. Heatmap \ref{fig:PCKMMonovsPCSKMeans_ARI} gives the general advantage to PCSKMeans for the ARI measure, but not by a wide margin, indicating no significant differences in some cases and advantage of PCKM-Mono in a significant portion of the experiments. When it comes to the comparison concerning NMI, and Unsat, conclusions remain unchanged. Heatmap \ref{fig:PCKMMonovsP2Clust_NMI} confirms the indisputable superiority of purely monotonic algorithms with respect to NMI. However, \ref{fig:PCKMMonovsPCSKMeans_Unsat} reveals no statistically significant differences between PCKM-Mono and PCSKMeans with respect to Unsat, and \ref{fig:PCKMMonovsCOPKMeans_Unsat} an advantage of PCKM-Mono over COP-Kmeans for the same measure. With this in mind, it is reasonable to assert that, for the experiments conducted in this study, the proposed PCKM-Mono algorithm has the same or better capabilities than previous CC algorithms to include constraints into the clustering process. Please note that, even if PCKM-Mono and PCSKMeans feature disputed results for the ARI measure, PCKM-Mono is indisputably superior to PCSKMeans for NMI and statistically similar to PCSKMeans regarding the Unsat measure, thus it is fair to claim an advantage of PCKM-Mono over PCSKMeans in the general case.

\section{A case of study: The Shanghai Ranking dataset} \label{sec:ShanghaiRank}

In this section we assess the applicability of our proposal for a real-world problem. The Shanghai Ranking of World Universities (SRWU) dataset has been used before to test the capabilities of monotonic clustering methods, e.g. in \cite{rosenfeld2021assessing} the top 100 institutions are used to test the P2Clust method. In our experiments we used the dataset available in \hyperlink{https://www.kaggle.com/code/saurav9786/eda-for-university-ranking/data?select=shanghaiData.csv}{this kaggle repository} \footnote{https://www.kaggle.com/code/saurav9786/eda-for-university-ranking/data?select=shanghaiData.csv}, which contains the SWRU results for years 2005 to 2015. Only the results from the year 2015 are used in our experiments. In our dataset, institutions are ranked from best to worst in chunks of size 50 for the first 100 institutions and in chunks of size 100 for the rest of them, generating a total of 7 classes for the 500 institution in the dataset. Our goal is to cluster the dataset so that institutions ranked in the same chunk appear in the same cluster in the final partition.

\begin{figure}[!ht]
	\centering
	\includegraphics[width=\linewidth]{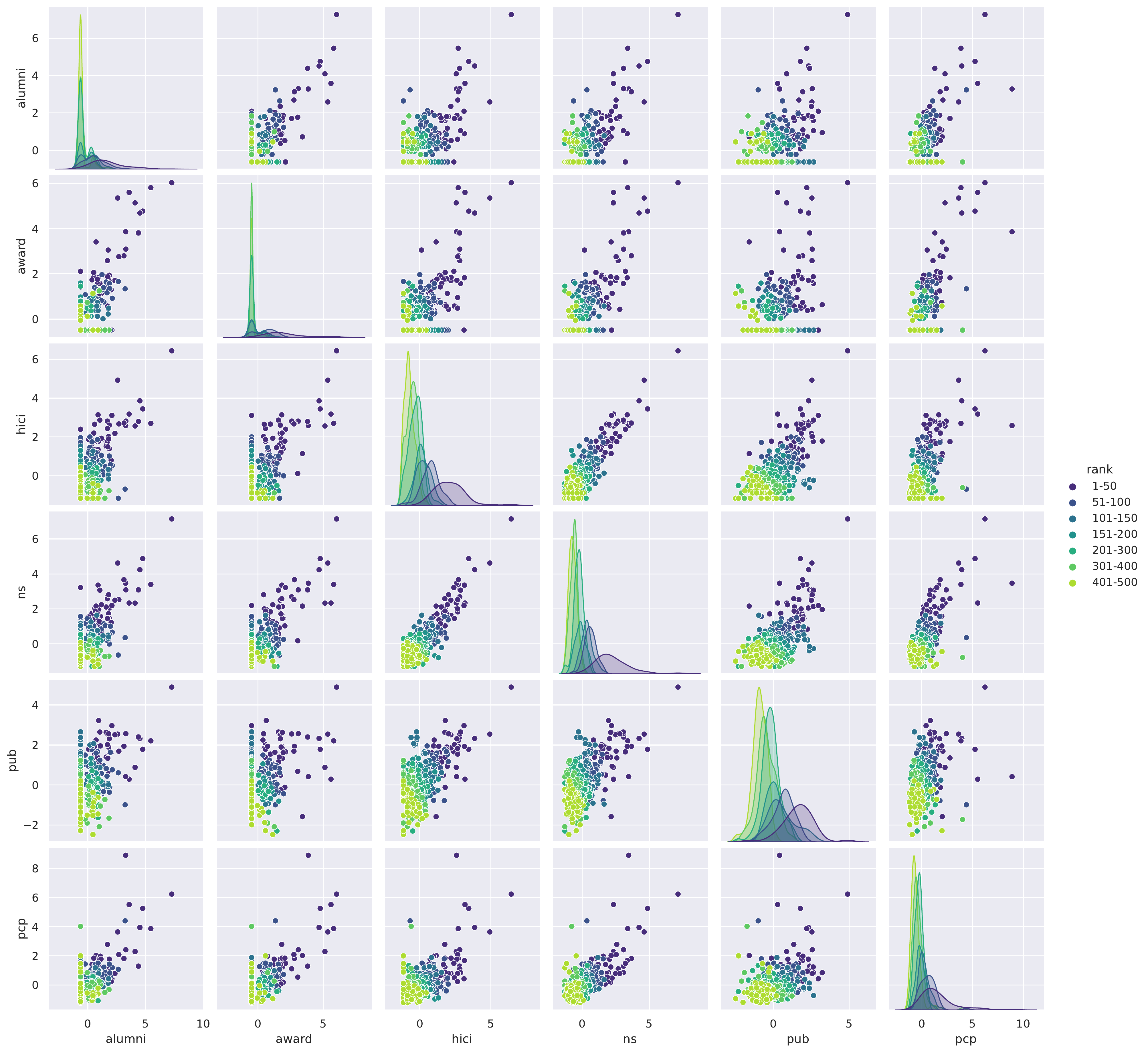}
	\caption{Pairplot for the Shanghai Ranking of world universities dataset}
	\label{fig:Pairplot}
\end{figure}

Originally, the dataset has 9 features, although some of them do not provide any valuable information for clustering methods and thus they can be removed, such as the institution name or its national rank. The dataset has 6 features after removing the useless ones, and can be visualized in the pairplot in Figure \ref{fig:Pairplot}. Observing Figure \ref{fig:Pairplot}, it seems clear that the SRWU is in fact a monotonic dataset, an therefore, it is has to be addressed with monotonic methods. However, there are some exceptions to this monotonicity. In fact, if we compute the NMI value for the true partition of the dataset, we obtain a value of $0.07$ as a result, which indicates that the monotonicity is broken by some instances. This is the reason why constraints can help improve the results, as if the dataset was purely monotonic, a method like P2Clust, which is hard constrained for monotonicity constraints, could solve it more accurately.

\begin{figure}[!h]
	\centering
	\subfloat[ ]{\includegraphics[width=0.5\linewidth]{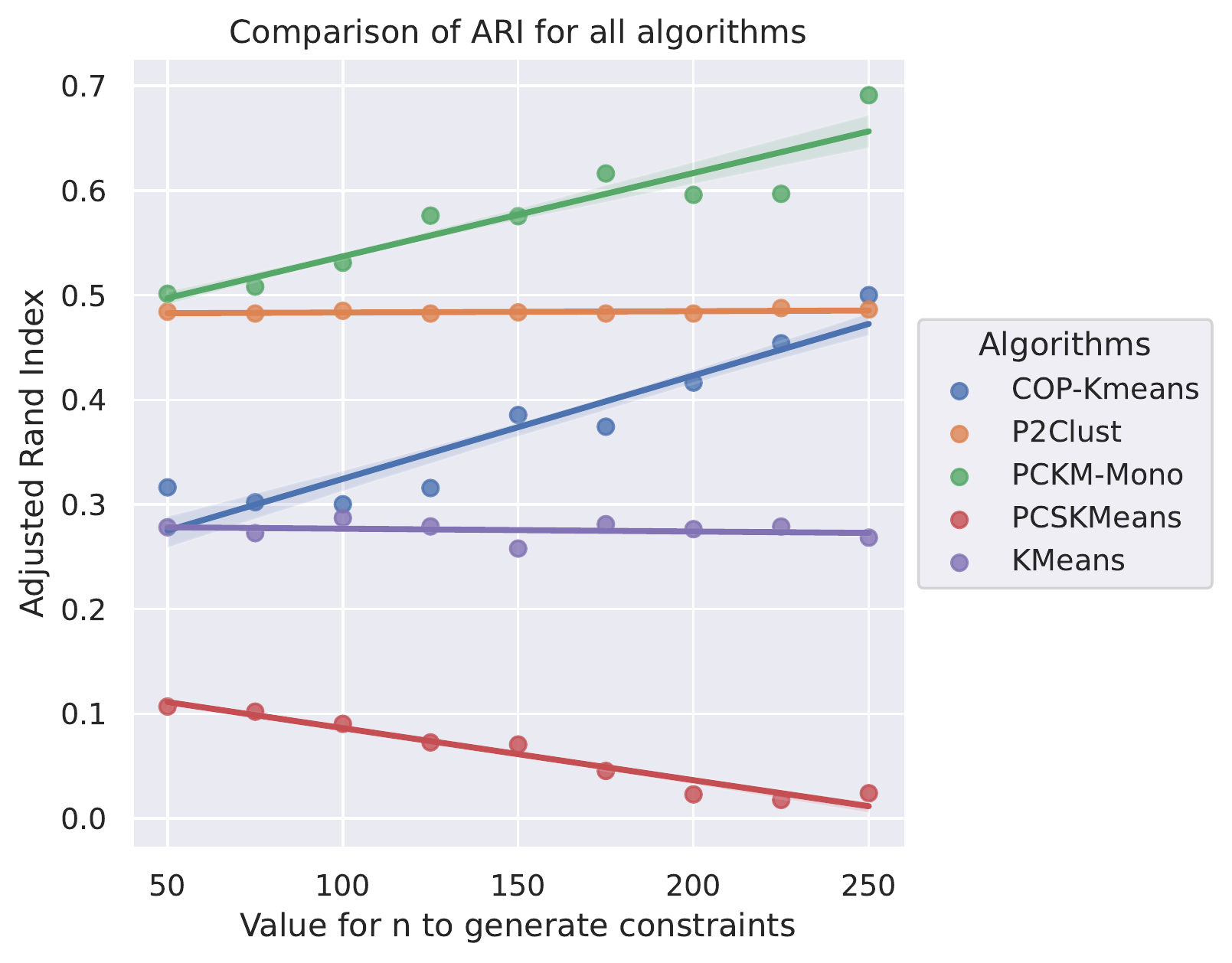}
		\label{fig:ARI_vs_constraints}}
	\subfloat[ ]{\includegraphics[width=0.5\linewidth]{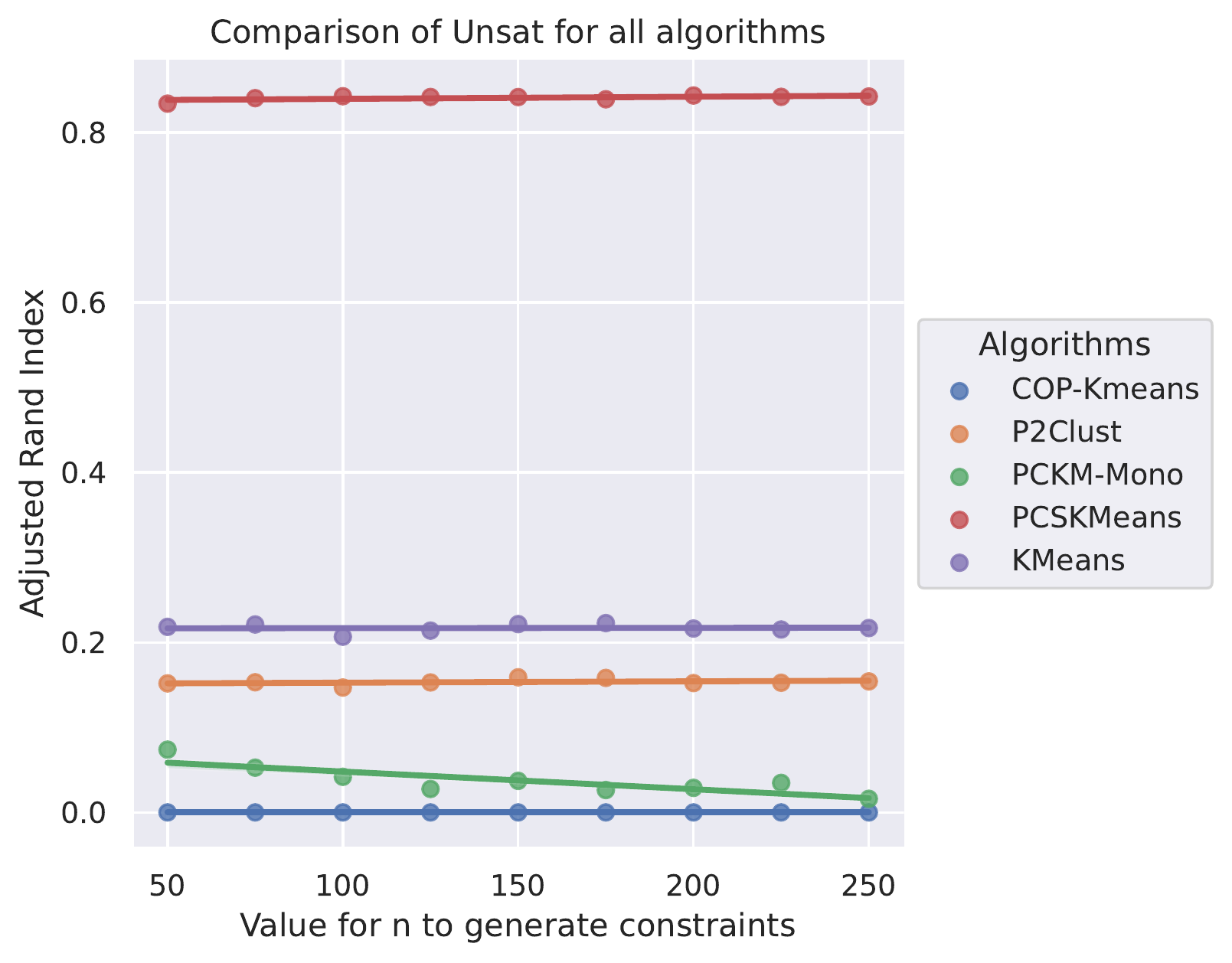}
		\label{fig:Unsat_vs_constraints}}
	\vspace{1.5\baselineskip}
	\subfloat[ ]{\includegraphics[width=0.5\linewidth]{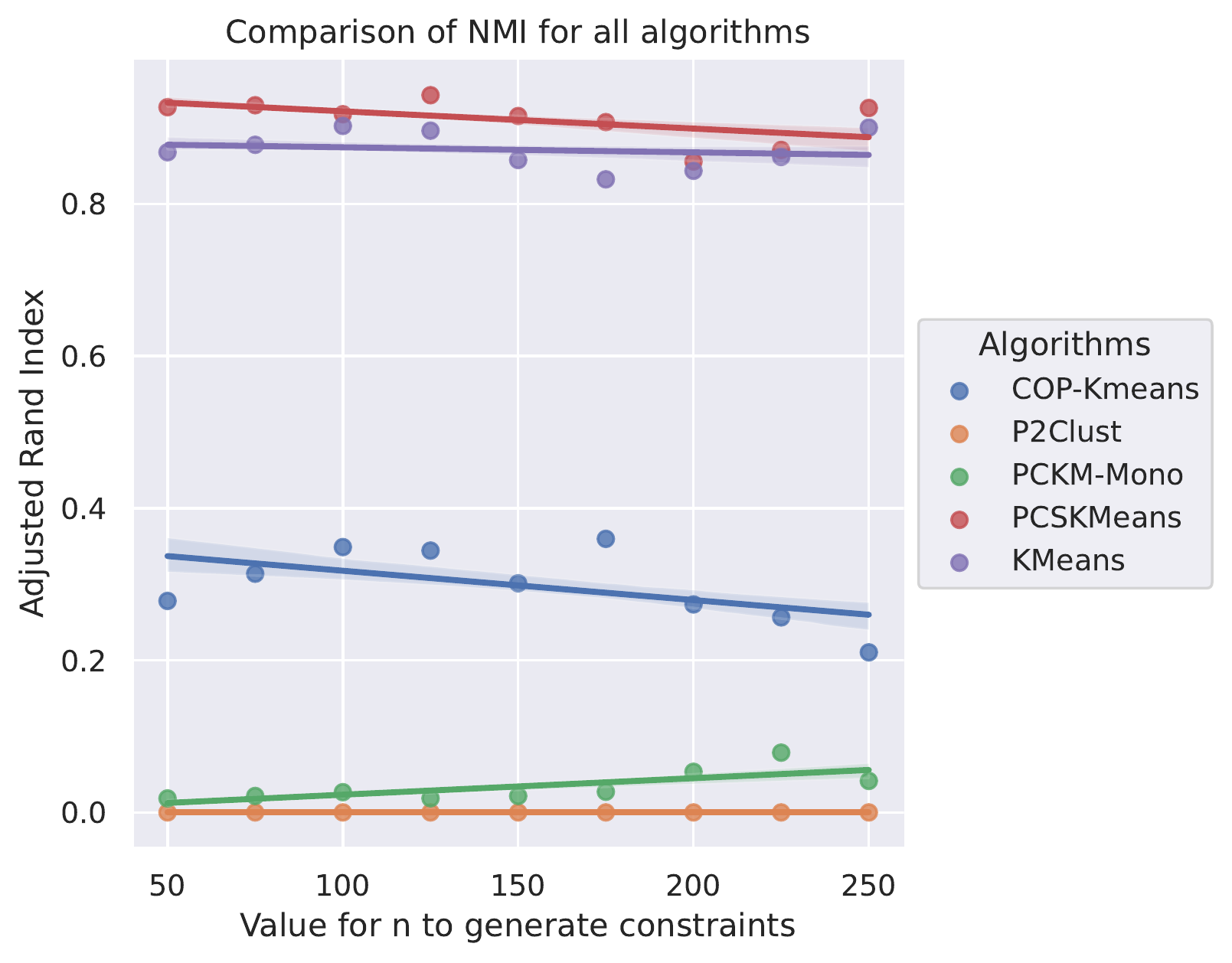}
		\label{fig:NMI_vs_constraints}}

	\caption{Scatterplot \ref{fig:ARI_vs_constraints} compares ARI results obtained by all five method compared in this study. Figures \ref{fig:Unsat_vs_constraints} and \ref{fig:NMI_vs_constraints} do so for the Unsat and NMI measures respectively.}
	\label{fig:SWRUresultsPlot}
\end{figure}

Scaling, standardization and missing values imputation (basic Knn imputer) steps are performed before applying all 5 clustering methods considered in this study to the dataset. Figure \ref{fig:SWRUresultsPlot} shows the results obtained by all method for the three quality measures and with increasing values for $n$ in the formula $(n_f(n_f-1))/2$, and thus, generating increasing levels of constraint-based information. This is conducted to observe how the results scale with the number of available constraints. Constraints are generated as it is done for benchmark datasets (see Section \ref{sec:ExpSetup}).

In Figure \ref{fig:ARI_vs_constraints} it can be clearly observed that PCKM-Mono represents the best option to generate scaling quality results for the SRWU partitioning problem. It is followed by the purely monotonic P2Clust method, which maintains stable results, as it does not consider constraints. It is also interesting to note how COP-Kmeans scale the results even by a greater factor than PCKM-Mono, although achieving worse results, as it cannot deal with the monotonicity of the data.

Regarding the results for Unsat, presented in Figure \ref{fig:Unsat_vs_constraints}, we can observe how Unsat values produced by PCKM-Mono scale inversely proportional with respect to the number of constraints. This is indicative of constraints helping the clustering process to find the true shape of the cluster, therefore making it easier for the method to satisfy a higher number of them. The rest of the methods maintain a stable Unsat, with COP-Kmeans always producing a value of 0 for this measure (as it can never generate partitions which violate any constraint) and with PCSKMeans featuring the worse value for it. This is indicative of the method not being suitable at all for the problem, as even non-constrained non-monotonic methods such as Kmeans are able to obtain better Unsat results.

In Figure \ref{fig:NMI_vs_constraints} we can observe one of the most interesting effects of constraints. Please note that NMI results for PCKM-Mono decline as the number of constraints increases. The interpretation of this result can be counterintuitive, as one could expect it to decrease. However, the NMI is actually shifting towards the NMI value produced by the true labels of SRWU ($0,07$), thus being more accurate in practice. With regard to non-constrained methods, they maintain an stable NMI value (as expected), with P2Clust always producing an NMI of 0, as it can never generate partitions which violate monotonicity. For the non-monotonic constrained clustering methods, it can be observed that the influence of constraints in COP-Kmeans is enough to divert clusters from the hyperspherical shape produced by the Euclidean distance, and thus generating an acceptable NMI value, which is not the case for PCSKMeans.

All of these results are in favor of the hypothesis of pairwise constraints and monotonicity constraints benefiting from each other when combined. Please note that, PCKM-Mono would produce the same NMI values as P2Cust if it were not for pairwise constraints, which have proved to divert the method from this trend and towards more accurate NMI results.

\section{Conclusion} \label{sec:Conclusions}

In this study, the first method which addresses Monotonic Constrained Clustering (MCC) is proposed:  Pairwise Constrained K-Means - Monotonic (PCKM-Mono). An expectation-minimization scheme is used to locally optimize a hybrid objective function, integrating a monotonic distance metric and a pairwise constraint-based penalty term. The experimental results obtained from a variety of datasets and their following statistical analysis confirm the viability of the proposed method when compared with purely monotonic and purely pairwise constrained clustering techniques. Even if PCKM-Mono obtains results similar to those obtained by previous approaches for specific monotonicity and pairwise constraint satisfaction, there is strong statistical evidence in favor of PCKM-Mono regarding general clustering quality measures.

\section*{Acknowledgements}

Our work has been supported by the research projects PID2020-119478GB-I00, A-TIC-434-UGR20 and PREDOC\_01648.

\section*{Conflict of interest}

The authors declare that there is no conflict of interest.

\clearpage

 \bibliographystyle{unsrt} 
 \bibliography{references}





\end{document}